\documentclass[letterpaper]{article} 
\usepackage{aaai23}  
\usepackage{times}  
\usepackage{helvet}  
\usepackage{courier}  
\usepackage{url}
\usepackage{graphicx} 
\usepackage{natbib}  
\usepackage{caption} 
\usepackage{microtype}
\usepackage{subfigure}
\usepackage{booktabs} 
\usepackage{algpseudocode}
\usepackage{amssymb}
\usepackage{epsfig}  
\usepackage{transparent}
\usepackage{algorithm}
\usepackage{amsmath,bm}
\usepackage{epstopdf, color}
\usepackage{mathrsfs} 
\usepackage{wrapfig}
\usepackage[english]{babel}
\usepackage{blindtext}
\usepackage{multicol}

\title{Generalizing LTL Instructions via Future Dependent Options}

\author{Duo Xu\thanks{Correspondance to: dxu301@gatech.edu}, Faramarz Fekri}

\affiliations{
  Department of Electrical and Computer Engineering\\
  Georgia Institute of Technology \\
  Atlanta, GA\\
}

\begin{document}

\maketitle

\begin{abstract}
    In many real-world applications of control system and robotics, linear temporal logic (LTL) is a widely-used task specification language which has a compositional grammar that naturally induces temporally extended behaviours across tasks, including conditionals and alternative realizations. An important problem in RL with LTL tasks is to learn task-conditioned policies which can zero-shot generalize to new LTL instructions not observed in the training. However, because symbolic observation is often lossy and LTL tasks can have long time horizon, previous works can suffer from issues such as training sampling inefficiency and infeasibility or sub-optimality of the found solutions. In order to tackle these issues, this paper proposes a novel multi-task RL algorithm with improved learning efficiency and optimality. To achieve the global optimality of task completion, we propose to learn options dependent on the future subgoals via a novel off-policy approach. In order to propagate the rewards of satisfying future subgoals back more efficiently, we propose to train a multi-step value function conditioned on the subgoal sequence which is updated with Monte Carlo estimates of multi-step discounted returns. In experiments on three different domains, we evaluate the LTL generalization capability of the agent trained by the proposed method, showing its advantage over previous representative methods. 
\end{abstract}

\section{Introduction}

Reinforcement learning (RL) is a promising framework for developing truly general agents capable of acting autonomously in the real world, ranging from video games \cite{mnih2015human,badia2020agent57} to robotics \cite{levine2016end,inala2021safe}.
Generalizing to different temporally extended tasks and following human instructions are key requirements for deploying autonomous agents in many real-world domains \cite{taylor2009transfer}. 
Linear temporal logic (LTL) \cite{pnueli1977temporal} is a popular means of specifying an objective for a reinforcement learning agent \cite{toro2018teaching,araki2021logical,vaezipoor2021ltl2action}. The compositional nature of tasks is reflected by the compositional and temporally-extended grammar of LTL specification. Instructions expressed in LTL encode temporal constraints that should be true during the comman execution. 
It is important for RL agent to learn to perform zero-shot execution of different LTL instructions by integrating the generalization abilities of deep learning models with the compositional structure of LTL. However, previous related works suffer from various shortcomings which can hinder real-world applications \cite{kuo2020encoding,araki2021logical,vaezipoor2021ltl2action,den2022reinforcement,liu2022skill}. Some works \cite{araki2021logical,den2022reinforcement,liu2022skill} solve new LTL instructions by leveraging the learned reusable skills or options, but produce sub-optimal or even infeasible solutions, especially when the symbolic observation is lossy, i.e., one symbolic state can correspond to multiple environment states.
These methods train an independent option for reaching a specific symbol or proposition as the subgoal, which can ignore the global optimality of task completion in the case of lossy symbolic observation. We have an example for this in Section \ref{sec:method}.
Further, \cite{kuo2020encoding,vaezipoor2021ltl2action} propose to train policies conditioned on the task formula directly, where the agent needs a large amount of environment samples to learn to understand temporal operators and figure out the optimal path for satisfying the formula. These approaches do not utilize reusable skills either. So, their sample efficiency of training is not satisfied in complex task or environment.

In this work, in order to tackle the above issues, we propose a novel multi-task RL approach for generalizing LTL tasks which can outperform previous methods in many aspects. 
We know the fact that every LTL formula can be decomposed into a list of subgoal sequences, any of which can satisfy the original formula \cite{leon2020systematic,leon2021nutshell}. Hence, in order to improve sample efficiency during the training, we train the agent to satisfy subgoal sequences instead of the original LTL formulas. Primarily we have two innovations.
1) We propose to learn options of satisfying subgoals conditioned on the sequence of future subgoals, taking the global optimality of task completion into consideration.
Specifically, the action value (Q) function and policy of every option are conditioned on the embedding of a sequence of future subgoals where the embedding is extracted by a GNN or GRU. Further, the option is trained not only with the experience of satisfying its subgoal, but also with the reward information of satisfying the future subgoal sequence.
2) Since the satisfaction of a subgoal sequence is temporally extended and can have long time horizon, in order to facilitate the reward propagation, we train a multi-step value function to predict the discounted return of satisfying a subgoal sequence. The innovation is that the value function is updated with Monte Carlo estimates of multi-step discounted return and, it sets the targets for updating the Q functions of options whenever a subgoal is satisfied. Hence, the reward information can be propagated throughout the state space in options more quickly and efficiently. In practical implementation, we also propose to use hindsight experience replay (HER) to relabel every unsuccessful trajectory, ensuring that there are enough trajectories to train the value function.

During testing, the unseen task formula $\varphi$ is first decomposed into a list consisting of all the sequences of subgoals that satisfy $\varphi$. Then the agent selects the best sequence $\xi$ of subgoals to execute which has the highest return predicted by the value function, considering both feasibility and optimality. This is because infeasible subgoal sequence can have very low predicted return. Finally, the agent adopts the trained options to satisfy subgoals in $\xi$ sequentially in the order fixed by $\xi$, so that the task $\varphi$ can be completed successfully. The safety is guaranteed with a high probability, by design, by avoiding the set of unsafe propositions which can falsify the task formula.


In experiments, we demonstrate the zero-shot generalization capability of the learned models in three environments, including both discrete and continuous domains. All these environments are procedurally generated where the layout and task specification are randomly generated, so that none of tasks here can be solved by simple tabular methods \cite{sutton2018reinforcement}. With comprehensive evaluations, we show that the proposed approach outperform previous representative methods in terms of sample efficiency, accuracy and optimality. 

\section{Related Work}
\label{sec:related}
Extending the RL paradigm to solve multiple temporal tasks has been studied by many previous works. These approaches augment the state space and obtain an equivalent product MDP by transforming the LTL formula into its automaton equivalence. Representative previous approaches, such as Q-learning for reward machines (Q-RM) \cite{camacho2019ltl,icarte2018using,icarte2022reward}, LPOPL \cite{toro2018teaching} and geometric LTL (G-LTL) \cite{littman2017environment}, augment the environment state space with the automaton transformation of the LTL specification. In addition, authors in \cite{jothimurugan2021compositional} proposed the DiRL framework to complete LTL task successfully by using hierarchical RL to interleave graph-based planning on the automaton and guide the agent's exploration for task satisfaction. However, although the compositional nature of LTL is utilized in these approaches to complete tasks, the compositionality is not leveraged in generalization to novel task specifications, so that the agent must learn the policy for satisfying a new LTL formula from scratch.

Learning independent option policies or skills for achieving each subgoal has been a common approach towards generalization in a temporal task setting for long  \cite{andreas2017modular,araki2021logical,leon2020systematic,leon2021nutshell}. For any unseen task formula, the agent sequentially composes these option policies to satisfy the task formula. However, a lot of additional fine-tuning is needed to satisfy the task formula correctly in these approaches, and they cannot address the issue of {\it lossy symbolic observation} so that the optimality and even feasibility of the solution can not be guaranteed. We propose a general framework for transferring learned policies to novel specifications in a zero-shot setting while preserving the ability to follow safety constraints.

Authors in \cite{kuo2020encoding} proposed learning a modular policy network by composing subnetworks via recurrent graph neural network for each proposition and operators, based on the syntax tree transformed from the LTL formula. Given a new task formula, the final policy network is created by composing the subnetwork modules in the new syntax tree corresponding to the given formula. Another paper \cite{vaezipoor2021ltl2action} proposes to use graph convolutional networks to learn an embedding for the given LTL formula to tackle novel LTL formulas. However, since the task formula is processed in its original form, the agent needs a lot of environmental interactions to learn to understand temporal operators and figure out the optimal path to satisfy the formula. These approaches may result in unsatisfactory performance on sample efficiency or optimality when the task formula has complex logic relationships. We compare these approaches with ours in experiments.

\section{Preliminaries}
\label{sec:prel}
\subsection{Reinforcement Learning}
RL provides a framework for learning to select actions in an environment in order to maximize the collected rewards over time \cite{sutton2018reinforcement}. RL deals with problems formalized as Markov decision processes (MDP). We here denote an MDP as a tuple $\mathcal{M}=\langle\mathcal{S}, \mathcal{A}, T, R, \gamma, {S}_0\rangle$, where $\mathcal{S}$ is a finite set of environment states, $\mathcal{A}$ is a finite set of agent actions, $T:\mathcal{S}\times\mathcal{A}\times\mathcal{S}\to[0,1]$ is a probabilistic transition function, $R:\mathcal{S}\times\mathcal{A}\to[R_{\text{min}}, R_{\text{max}}]$ is a reward function with $R_{\text{min}}, R_{\text{max}}\in\mathbb{R}$, $\gamma\in[0,1)$ is a discount factor, $S_0:s_0\sim S_0$ is a distribution of initial states. In each time step $t$, the agent observes the environment state $s_t$ and selects an action $a_t$ to apply, according to a policy function $\pi\in\Pi:\mathcal{S}\times\mathcal{A}\to[0,1]$, and then collects reward $r_t=R(s_t,a_t)$. 

For some policy $\pi$, the values V and Q for any state $s$ and state-action pair $(s,a)$ at time $t$ can be defined as below,
\begin{eqnarray}
    V_{\pi}(s)&=&\mathbb{E}_{\pi}\bigg[\sum_{\tau=t}^{\infty}\gamma^{\tau-t}r_{\tau}|s_t=s\bigg], \nonumber \\ 
    Q_{\pi}(s,a)&=&\mathbb{E}_{\pi}\bigg[\sum_{\tau=t}^{\infty}\gamma^{\tau-t}r_{\tau}|s_t=s, a_t=a\bigg] \label{critic}
\end{eqnarray}
where $\mathbb{E}_{\pi}$ is the expectation of accumulated rewards following some policy $\pi$. A policy is the optimal policy $\pi^*$ if it produces the highest accumulated rewards: $\forall s\in\mathcal{S}, \forall\pi\in\Pi, \forall a\in\mathcal{A}: Q_{\pi^*}(s,a)>Q_{\pi}(s,a)$. Searching $\pi^*$ can be addressed by parameterizing the policy and finding optimal parameters $\theta^*$ that maximize the accumulated rewards by a learning algorithm. Specifically, parameters $\theta$ can be weights of neural networks optimized by gradient descent. 

A widely-used parameterized approach of searching $\pi^*$ in the space of neural networks is known as deep Q-Networks (DQN) \cite{mnih2015human}. DQN uses deep neural networks with weights $\theta$ to approximate $Q_{\pi}(s,a|\theta)$. Then, at each step $t$, the agent selects actions uniform randomly with some probability $\epsilon\in[0,1)$ or greedily over $Q_{\pi}(s,a)$ with probability $1-\epsilon$. The generated experience tuple $(s_t, a_t, r_t, s_{t+1})$ is stored to a buffer $\mathcal{B}$. The weights $\theta$ are updated by using advanced optimizer, such as Adam, iteratively. At each iteration, we update the weights $\theta$ of neural networks by minimizing the loss function as below
\begin{equation}
    \mathcal{L}(\theta;\theta^-)=\mathbb{E}_{(s,a,r,s')\sim\mathcal{B}}\bigg(r+\gamma\max_{a'} Q(s',a'|\theta^-)-Q(s,a|\theta)\bigg)^2
    \label{q-learning}
\end{equation}
where $\theta^-$ are target weights of neural networks which are updated periodically for improving numerical stability of the learning process.

\subsection{Option Framework}
\label{sec:pre_option}
The option framework was introduced in \cite{sutton1999between} to incorporate temporally-extended actions (options) into reinforcement learning. An option $o=\langle\mathcal{I}, \beta, \pi\rangle$ is defined by three elements: 1) the initiation set $\mathcal{I}$ denotes the states where the option can be started to execute; 2) the termination condition $\beta$ defines the condition when option execution ends; 3) the option policy $\pi$ selects actions to take the agent to realize $\beta$ starting from any state in $\mathcal{I}$. We leverage the options framework to define the task-agnostic skills in LTL generalization.

\subsection{Linear Temporal Logic (LTL)}
\label{sec:ltl}
An LTL formula $\varphi$ is a boolean function that determines whether the objective formula is satisfied by the given trajectory or not \cite{pnueli1977temporal}. When formulating the task formulas for the RL agent, the first step is to specify a common vocabulary shared by both the environment and the agent. In this work, a finite set of propositions (symbols) $\mathcal{P}$ is used as the vocabulary, representing high-level events or properties of the environments. There is a labelling function for detecting the occurrences of these propositions in the environment. For instance, in service robot environment, $\mathcal{P}$ could include events such as opening the drawer, activating the fan, turning on/off the stove, or entering the bathroom. Then, by using LTL, the task of the agent can also include temporally-extended occurrences of these events. For example, two possible tasks that can be expressed in LTL are (1) "Open the drawer and activate the fan in any order, then turn on the stove" and (2) "Open the drawer but do not enter the bathroom until the stove is turned off". 

Given a finite set of propositions $\mathcal{P}$, the grammar of an LTL formula is expressed as below:
\begin{equation}
    \varphi::=p|\neg\varphi|\varphi\vee\phi|\bigcirc\varphi|\varphi\cup\phi\hspace{10pt}\hspace{15pt}\forall p\in\mathcal{P} \nonumber
\end{equation}
Since it is defined over temporally-extended events, we use sequences of symbolic observations (i.e., mapping from the observed state to a set of propositions in $\mathcal{P}$) to evaluate LTL formulas. Specifically, the operators $\neg$ (not), $\vee$ (or), $\wedge$ (and) are same as propositional logic operators. The formula $\bigcirc\varphi$ (next $\varphi$) means that $\varphi$ should be satisfied at next time step, and $\varphi\cup\phi$ ($\varphi$ until $\phi$) means that $\varphi$ should hold until $\phi$ is satisfied.

\begin{figure}
    \centering
    \includegraphics[width=1.5in]{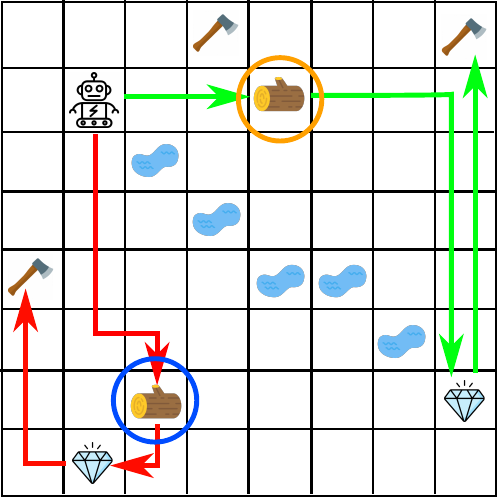}
    \caption{Motivating example. The LTL task is $\Diamond(\text{wood}\wedge\Diamond(\text{diamond}\wedge\Diamond\text{ax}))$ (go to collect wood, then diamond and finally ax). The wood in orange circle denotes the state $s_A$ and that in blue circle denotes state $s_B$.}
    \label{fig:motivating}
    \vspace{-10pt}
\end{figure}

{\it The progression function} takes an LTL formula and the current labelled state (symbolic observation) as inputs and returns a formula that identifies aspects of the original formula that remain to be addressed \cite{bacchus2000using,vaezipoor2021ltl2action}. Specifically, for any LTL formula $\varphi$ and a truth assignment $\sigma$ over $\mathcal{P}$, the progression function in terms of $\sigma$ and $\varphi$ is defined as $\text{prog}(\sigma, \varphi)$. It is semantics-preserving, since the progress towards completion of the task is reflected in the remaining formulas. For example, in Figure \ref{fig:motivating}, consider task $\varphi:=\Diamond(\text{wood}\wedge\Diamond\text{diamond})$ (collect wood and then diamond), which will progress to $\Diamond\text{diamond}$ as soon as the agent collects wood. 

In this work, LTL progression has two usages during testing. First, given any task described by LTL formula $\varphi$, $\varphi$ is progressed by the observed symbolic state ($L(s_t)$) in every time step, making the reward dependent on task satisfaction Markovian \cite{icarte2018using,vaezipoor2021ltl2action}. 
The other usage of LTL progression is to predict propositions which can falsify the task $\varphi$ and have to be avoided by the agent, so that the task can be finished safely. 

\subsection{RL with LTL Tasks}
\label{sec:rl_ltl}
Assume that the agent is working on an environment MDP $\mathcal{M}_e=\langle\mathcal{S}, \mathcal{A}, T, R_e, \gamma, S_0\rangle$, a labelling function $L:\mathcal{S}\times\mathcal{A}\to2^{\mathcal{P}}$, a finite set of LTL formulas $\Phi$, a probability distribution $\tau$ over formulas $\varphi$ in $\Phi$, our target is to learn a multi-task agent with policy $\pi(a|s, \varphi)$ which can finish the task $\varphi$ by maximizing both the environment reward $R_e$ and the task satisfaction reward $R_{\varphi}$. The episode ends when the task is completed, falsified, or a terminal state is reached. All of these can be formulated as a taskable MDP \cite{illanes2020symbolic,vaezipoor2021ltl2action} as below.



\noindent
{\bf Definition 1.} Given an environment MDP $\mathcal{M}_e=\langle\mathcal{S}, \mathcal{A}, T, R_e, \gamma, S_0\rangle$, a finite set of propositions $\mathcal{P}$, a labelling function $L:\mathcal{S}\times\mathcal{A}\to2^{\mathcal{P}}$, a finite set of LTL formulas $\Phi$, and a probability distribution $\tau$ over $\Phi$, we construct taskable MDP as $\mathcal{M}_{\Phi}=\langle\mathcal{S}',\mathcal{A},T',R',S_0'\rangle$, where $\mathcal{S}'=\mathcal{S}\times\text{cl}(\Phi)$, $T'(s',\varphi'|s,\varphi,a)=T(s'|s,a)$ if $\varphi'=\text{prog}(L(s,a),\varphi)$ (zero otherwise), $S_0'(s,\varphi)=S_0(s)\cdot\tau(\varphi)$, and
\begin{equation}
    R'(\langle s, \varphi\rangle, a)=\begin{cases}
    R_F\hspace{18pt}\text{if }\hspace{3pt}\text{prog}(L(s,a), \varphi)=\text{true} \\
    -R_F\hspace{11pt}\text{if }\hspace{3pt}\text{prog}(L(s,a), \varphi)=\text{false} \\
    R_e\hspace{18pt}\text{otherwise}
    \end{cases} \nonumber 
\end{equation}
{\bf LTL Satisfaction in Finite Steps.} We have to make sure that the agent can receive signals of task satisfaction or falsification in finite steps. The episode terminates when LTL formula $\varphi$ is satisfied or falsified, or the maximum length of a episode is exceeded. Since every episode has finite length, we have to determine the LTL formula to be satisfied or unsatisfied in a finite number of steps. This is guaranteed for the case of co-safe LTL \cite{kupferman2001model} in which $\Box$ (always) is not allowed and $\bigcirc, \bigcup$ and $\Diamond$ are only used in the positive normal form. For those LTL formulas which cannot be verified or falsified in finite time (e.g. $\Box\neg\text{lake}$), we alter the reward function to render an appropriate reward ($R_F/-R_F$) after a very large but finite number of steps (i.e., maximum length of an episode). For instance, if the formula $\Box\neg\text{lake}$ is not falsified at the end of an episode, it is regarded being satisfied and the agent can get a positive reward ($R_F$).


\section{Methodology}
\label{sec:method}
In contrast to previous papers on multi-task RL with LTL instructions, we improve the learning performance of the agent by addressing some important practical issues here. The first issue is {\it the lossy symbolic observation},
meaning that a single propositional symbol can correspond to multiple different environment states, e.g., in Figure \ref{fig:motivating} the proposition "wood" corresponds to two different states where the agent reaches the wood in the second row (state $s_A$) or the second last row (state $s_B$). The agent should be able to choose the best subgoal state to reach by considering the global optimality of task completion. 
The second issue is that, the rewards for satisfying subgoal sequence can be difficult to propagate throughout the state space via Q functions of options. This is because the subgoal sequence consists of temporally extended subgoals with long horizons, and Q functions of options are updated by the one-step temporal difference (TD-1) method \cite{sutton2018reinforcement}.
These issues are common in real-world problems but ignored by previous LTL-RL works \cite{andreas2017modular,leon2020systematic,araki2021logical,leon2021nutshell,vaezipoor2021ltl2action}.

\noindent
{\bf Motivating Example. } In Figure \ref{fig:motivating}, assume that environment reward $R_e$ is $-0.1$ for every movement and the given task is $\varphi:=\Diamond(\text{wood}\wedge\Diamond(\text{diamond}\wedge\Diamond\text{ax}))$ (go to collect wood, then diamond and finally ax). There are two choices ($s_A$ and $s_B$) for the agent to collect wood. Previous option-based approaches may myopically choose to collect the wood in the second row which is closer, and finish the task $\varphi$ along the green path. However, considering $R_e$, the globally optimal solution of task $\varphi$ is the red path. 
In some cases, the decision made by myopic option-based approaches may lead to infeasible solutions. For instance, when the game in Figure \ref{fig:motivating} has constraint that the agent cannot move more than 12 steps in one episode, the green path with myopic choice of collecting wood is infeasible.

In this work, in order to address issues mentioned above, we propose a novel option framework where options are dependent on the sequence of future subgoals. Let $o_p^{\xi}$ denote the option of reaching subgoal $p$ conditioned on $\xi$ as a sequence of future subgoals to satisfy. We train each option $o_p^{\xi}$ not only by the experience of reaching the subgoal $p$, but also with the reward information of satisfying subgoals in $\xi$ (in a fixed order same as $\xi$). 
Further, in order to facilitate the reward propagation in long-horizon tasks, such as sequence $\xi$ of future subgoals, we also train a multi-step value function $V^{\phi}(s;\xi)$ to predict the discounted return obtained by reaching subgoals in sequence $\xi$ starting from the state $s$, which is updated by the Monte Carlo estimates of multi-step discounted return. We use $V^{\phi}$ to set target values to update Q functions of options, hence accelerating the reward propagation in options.

Going back to the motivating example in Figure \ref{fig:motivating}, when the option of collecting "wood" is also trained with reward information of collecting diamond and then ax after collecting wood, i.e., $p=$"wood" and $\xi:=$["diamond", "ax"], the trained $\pi_{p}^{\xi}$ will make the agent to choose the wood in the last 2nd row ($s_B$) instead of the upper one ($s_A$). This is because $V^{\phi}(s_A;\xi)<V^{\phi}(s_B;\xi)$ and $V^{\phi}$ sets the targets for updating Q function of $\pi_{p}^{\xi}$. 

In training, the option policies of the agent are trained to satisfy a randomly generated subgoal sequence in a procedurally generated environment, and their corresponding Q and value functions are trained with agent's experience in the replay buffer by an off-policy RL algorithm.
In the testing, the unseen LTL task described by formula $\varphi$ is solved by the agent following three steps as below without further learning:
\begin{enumerate}
    \item Decompose the task formula $\varphi$ into a list of subgoal sequences (or paths), i.e., $\mathcal{K}:=\{\tau_i\}_{i=1}^{M_{\varphi}}=\{[p^i_1, p^i_2, \ldots, p^i_{L_i}]\}_{i=1}^{M_{\varphi}}$ such that $\varphi$ can be satisfied by every $\tau_i$. For instance, if $\varphi=\Diamond(a\wedge\Diamond((b\vee c)\wedge\Diamond(d\vee e)))$, the decomposed subgoal sequences are $\mathcal{K}=\{[a,b,d], [a,b,e], [a,c,d], [a,c,e]\}$ and $M_{\varphi}=4$; 
    \item Select the optimal sequence $\tau^*$ from $\mathcal{K}$ based on the value function $V^{\phi}$; 
    \item Use corresponding options to reach every subgoal with future subgoals in $\tau^*$, so that task formula $\varphi$ can be satisfied with the global optimality considered.
\end{enumerate}

\begin{figure*}
    \centering
    \fontsize{9pt}{10pt}\selectfont
    \def\svgwidth{4.8in}
    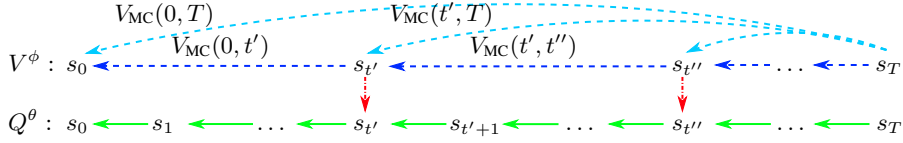
    \caption{Diagram of back-propagation of reward information. The green line shows that Q functions $Q^{\theta}$ of options are learned by TD-1 method. The first and second subgoals ($\xi[0]$ and $\xi[1]$) are satisfied at $s_{t'}$ and $s_{t''}$, respectively. The red line shows that the value function $V^{\phi}$ sets the target value for $Q^{\theta}$ whenever a subgoal is satisfied. The blue and cyan curves denote Monte Carlo of multi-step discounted return $V_{\text{MC}}(\cdot,\cdot)$ with range of time steps denoted. The value function $V^{\phi}$ is updated with the same pace as subgoal satisfaction, which has much coarser time resolution than $Q^{\theta}$.}
    \label{fig:multi-v}
    \vspace*{-10pt}
\end{figure*}

\subsection{Future Dependent Option}
\label{sec:option}
In this work, we define a future dependent option, i.e., $o_p^{\xi}:=\langle\mathcal{S}, \beta_p, \pi_p^{\xi}\rangle$ where $\xi$ is a finite sequence of subgoals to be satisfied in the future after $p$. We regard every proposition $p\in\mathcal{P}$ in taskable MDP (Definition 1 in Section \ref{sec:rl_ltl}) as the subgoal for an option to reach. Specifically, without loss of generality, the initial set is the same as the state space $\mathcal{S}$, and the terminal function is the indicator of satisfying subgoal $p$, i.e., $\beta_p(s)=\bm{1}\{L(s)\models p\}$ where $L(\cdot)$ is the labeling function defined in Section \ref{sec:rl_ltl}. The option policy $\pi_p^{\xi}$ is trained to maximize the discounted return of satisfying $p$ as subgoal. Additionally, the reward information of satisfying $\xi$ is also back-propagated to train the action value (Q) function of $\pi_p^{\xi}$ via the value function $V^{\phi}$, encouraging the option policy to achieve the global optimality of satisfying both $p$ and $\xi$. 

To improve the sample efficiency, we use off-policy RL method to train every option. Generally, for any option policy $\pi_p^{\xi}$, the agent learns a sample-based approximation to the action value (Q) function $Q_{\pi_{p}^{\xi}}(s,a)$ in \eqref{critic}, denoted as $Q_{p}^{\theta}(s,a;\xi)$. The Q function of option is updated by TD-1 method as \eqref{q-learning}. When the action is discrete, $\pi_p^{\xi}$ can be directly induced from $Q_{p}^{\theta}(s,\cdot;\xi)$ (the action with highest Q value). When the action is continuous, we need to learn an actor network for $\pi_p^{\xi}$ to approximate the maximizer (action) of $Q_{p}^{\theta}(s,\cdot;\xi)$ via the SAC algorithm \cite{haarnoja2018soft}.

The option policy $\pi_p^{\xi}$ is trained together with other options used to satisfy subgoals in $\xi$. 
Specifically, given any sequence of subgoals $\kappa:=\{p_i\}_{i=1}^K$ for training, denoting $\xi_k:=\{p_i\}_{i=k+1}^K$ and $k=1,\ldots,K$, we execute option policies starting from $\pi_{p_1}^{\xi_1}$, and when the subgoal $p_k$ in $\kappa$ is satisfied, we switch to $\pi_{p_{k+1}}^{\xi_{k+1}}$, until the agent satisfies the last subgoal $p_K$ by using policy $\pi_{p_K}^{\varnothing}$ ($\varnothing$ denotes empty). In addition to environmental rewards, the agent will receive the reward $R_F$ (defined in Section \ref{sec:rl_ltl}) when the last subgoal $p_K$ is satisfied. For any $k=1,\ldots,K-1$, the discounted return during the execution of options from $o_{p_{k+1}}^{\xi_{k+1}}$ to $o_{p_K}^{\varnothing}$ are all back-propagated to train the policy $o_{p_k}^{\xi_k}$ (updating $Q^{\theta}_p(\cdot,\cdot;\xi_k)$), via the value function $V^{\phi}$. 

\subsection{Multi-step Value Function}
\label{sec:value}
Since the Q functions of option policies are updated with TD-1 method in \eqref{q-learning}, each update can propagate the reward information for only one time step. However, we note that the satisfaction of a subgoal sequence $\xi$ can have long horizon and sparse rewards. Therefore, it can be inefficient to propagate the reward information of satisfying $\xi$ back to train $Q^{\theta}_p(\cdot,\cdot;\xi)$ via lagged Q network as \eqref{q-learning} (TD-1 method). In the rest of the paper, we use $\xi[k]$ to denote the $k$-th subgoal in sequence $\xi$.

In order to tackle this issue, we propose to learn a multi-step value function $V^{\phi}(s;\xi)$ to estimate the discounted return of satisfying the subgoal sequence $\xi$ starting from state $s$. It is used to set the target value for updating $Q^{\theta}_p(\cdot,\cdot;\xi)$ so that the reward propagation toward option Q functions can be accelerated. In Figure \ref{fig:multi-v}, it shows how the reward information is back-propagated in both $Q^{\theta}$ and $V^{\phi}$ visually.

Specifically, the target value for updating $V^{\phi}$ is calculated based on Monte Carlo estimates of two discounted returns. The first is the Monte Carlo estimate of the discounted return till the end of the trajectory (cyan curves in Figure \ref{fig:multi-v}),
i.e., $V_{\text{MC}}(t, T):=\sum_{k=t}^T\gamma^{k-t}r_t$ ($T$ is the last time step of the trajectory).
We use $V_{\text{MC}}(t, T)$ here since it is unbiased and good at capturing long-term rewards, but it also has large variance \cite{sutton2018reinforcement}. Then we also calculate another Monte Carlo estimate of discounted return till the satisfaction of next subgoal $\xi[0]$ (blue curves in Figure \ref{fig:multi-v}), i.e., $V_{\text{MC}}(t, t')=\sum_{k=t}^{t'}\gamma^{k-t}r_t$ ($t'$ is the time when $\xi[0]$ is satisfied). This $V_{\text{MC}}(t,t')$ is used to build a multi-step temporal difference (TD) target for updating $V^{\phi}$, together with the value estimate of satisfying other subgoals $\xi[1:]$ from a lagged value network $V^{\phi^-}$ \cite{van2016deep}. 

Assume we have a trajectory $\tau=\{s_0,a_0,r_0,s_1,\ldots,s_{T-1},a_{T-1},r_{T-1},s_T\}$ with subgoal sequence $\xi$ as the task to finish. 
If next subgoal $\xi[0]$ is satisfied at time $t'$, the target of value function is written as 
\begin{eqnarray}
    \lefteqn{V^{\text{target}}(s_t;\xi)=\sum_{k=t}^{t'}\gamma^{k-t}r_t} \nonumber \\ 
    && + \gamma^{t'-t}\max\{V^{\phi^-}(s_{t'};\xi[1:]), V_{\text{MC}}(t', T)\} \label{v-target}
\end{eqnarray}
In the equation above, the first term is $V_{\text{MC}}(t,t')$ which forms a multi-step TD target together with a lagged value network $V^{\phi^-}$. As discussed above, we also use $V_{\text{MC}}(t',T)$ in \eqref{v-target}. Since value network always has very small values throughout the state space in early training, we need to use a maximum operator in \eqref{v-target} to help the reward propagate from the end of the trajectory.
If next subgoal $\xi[0]$ is not satisfied by any state in $\tau$, the target will become $V^{\text{target}}(s_t;\xi)=\max\{V^{\phi^-}(s_t;\xi), V_{\text{MC}}(t, T)\}$ to facilitate the reward propagation. Finally, the value function $V^{\phi}$ is trained to predict its target value by optimizing the loss function
\begin{equation}
    J(\phi)=\ell(V^{\phi}(s_t;\xi), V^{\text{target}}(s_t;\xi)) \label{v-train}
\end{equation}
where $\ell$ is an arbitrary differentiable loss function.

\begin{figure*}
    \centering
    \subfigure[Letter]{
        \centering
        \fontsize{6pt}{10pt}\selectfont
        \def\svgwidth{1.in}
        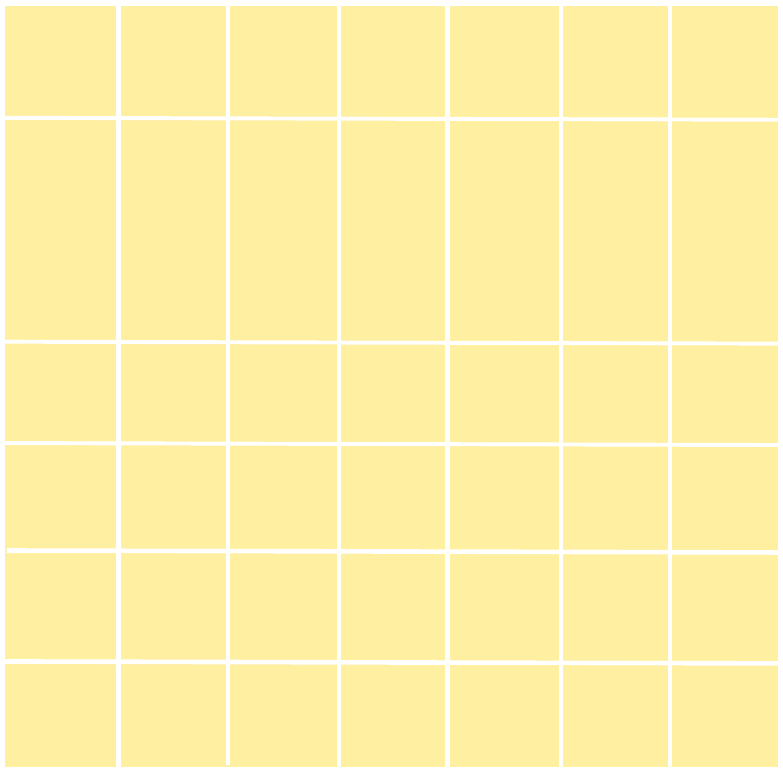
        \label{fig:fsa2}
    }
    \hspace{5pt}
    \subfigure[Room]{
        \centering
        \includegraphics[width=1.in, height=1.in]{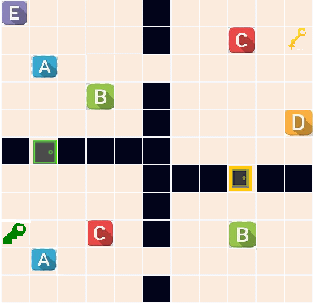}
        \label{fig:room}
    }
    \hspace{5pt}
    \subfigure[Navigation]{
        \centering
        \includegraphics[width=1.in, height=1.in]{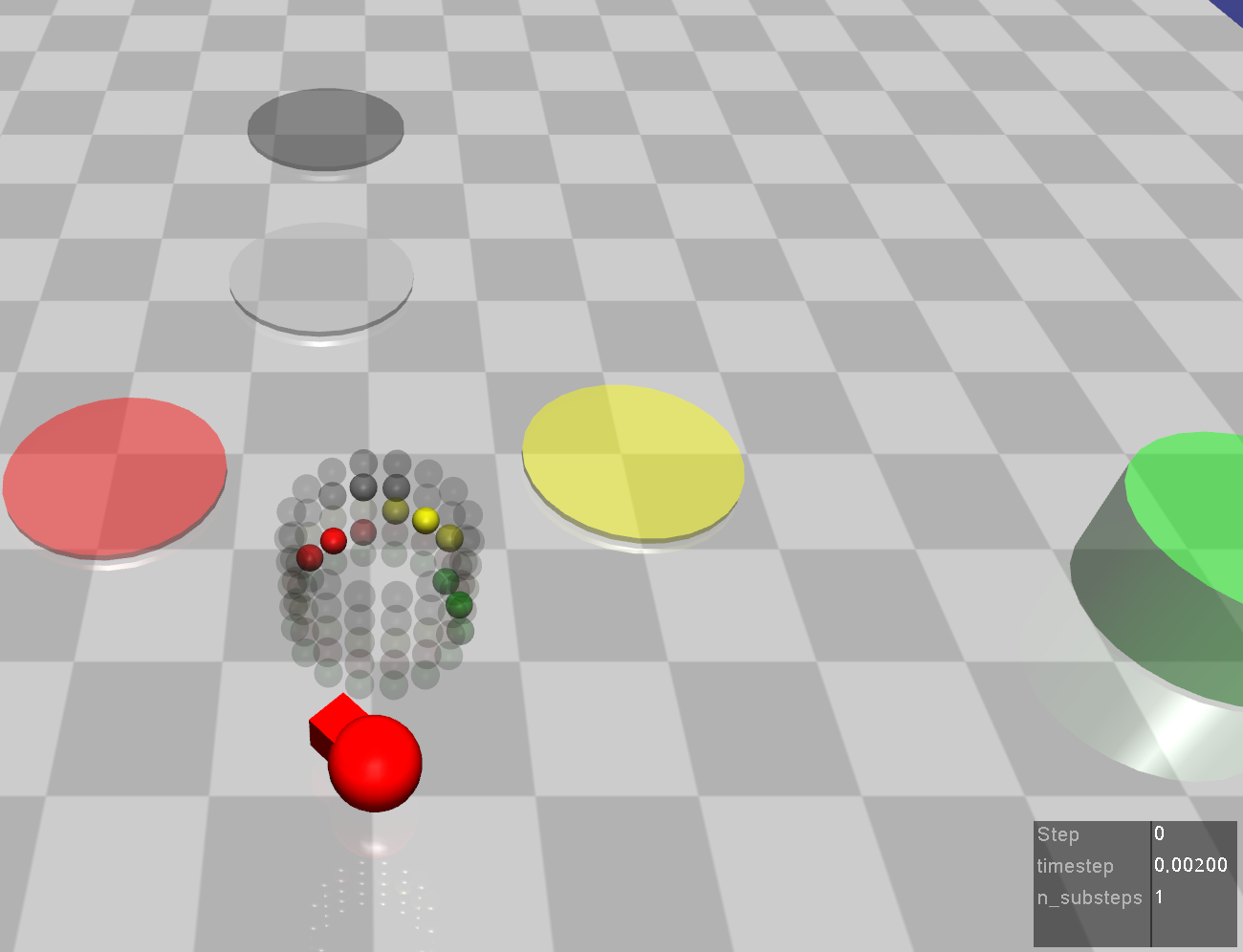}
        \label{fig:robot}
    }
    \vspace*{-5pt}
    \caption{Environments. Note that these environments are procedurally generated and hence tasks cannot be solved by simple tabular methods.}
    \vspace*{-10pt}
\end{figure*}

The value function $V^{\phi}$ sets the target value to update the Q functions of option policies $Q_p^{\theta}(\cdot,\cdot;\xi)$ whenever a subgoal is satisfied. 
For any tuple $(s_t,a_t,r_t,s_{t+1})$, the target value for $Q_p^{\theta}(\cdot,\cdot;\xi)$ is expressed as,
\begin{eqnarray}
\lefteqn{Q_p^{\text{target}}(s_t,a_t,r_t,s_{t+1};\xi) = r_t+ \gamma\beta_p(s_{t+1}) V^{\phi}(s_{t+1};\xi)} \nonumber \\
&& + \gamma(1-\beta_p(s_{t+1}))\max_{a'} Q_p^{{\theta}^-}(s_{t+1},a';\xi) \label{q-target}
\end{eqnarray}
where ${\theta}^-$ is the parameter of the lagged target network as \cite{van2016deep}. This target means that when $p$ is not satisfied yet ($\beta_p(s_{t+1})=$false), the Q function is updated via the classical TD-1 method. However, whenever $p$ is satisfied ($\beta_p(s_{t+1})=$true), it is updated with the target value given by $V^{\phi}(\cdot;\xi)$ which can quickly propagate discounted return of satisfying $\xi$ back to $s_{t+1}$, achieving the global optimality of satisfying both $p$ and $\xi$. Then $Q_p^{\theta}(\cdot,\cdot;\xi)$ can be updated by minimizing the loss as below,
\begin{equation}
    J(\theta)=\mathbb{E}_{(s,a,r,s',p,\xi)\sim\mathcal{B}}\big[\ell(Q_p^{\theta}(s,a;\xi), Q_p^{\text{target}}(s,a,r,s';\xi))\big] \label{q-train}
\end{equation}
where $\mathcal{B}$ is the replay buffer. 

\subsection{Practical Implementation}
\label{sec:implement}
The complete implementation techniques are introduced in Appendix \ref{sec:practical2}. The algorithms for training and testing are presented in Algorithm \ref{alg:training} and \ref{alg:testing} in Appendix \ref{sec:algos}. 


\vspace{5pt}
\noindent
{\bf Hindsight Experience Replay.} In early learning stage, most trajectories produced by agent's policies cannot achieve or satisfy the given task, which cannot provide any useful reward information to train agent's policy and value functions.
Therefore, in the training of the multi-task agent we propose to modify the hindsight experience replay (HER) \cite{andrychowicz2017hindsight} to better utilize the past trajectories of the agent and hence improve the learning efficiency. 
Specifically, HER is extended to temporal logic domain by modifying any unsuccessful trajectory whose given task was not successfully finished. Therefore, in any unsuccessful trajectory $\tau$ with subgoal sequence $\xi$ ($\xi$ is not finished by $\tau$), we find $\xi'$ which is the subgoal sequence satisfied by $\tau$ actually and replace $\xi$ by $\xi'$, so that the trajectory $\tau$ with task $\xi'$ becomes a successful trajectory ($\xi'$ is finished by $\tau$). Then, assigning a large positive reward $R_F$ at the time step when $\xi'[-1]$ becomes satisfied makes the trajectory $\tau$ useful to the training of options and value function.


\section{Experiments}
\label{sec:experiments}
Our experiments are designed to evaluate the performance of multi-task RL agent trained by the proposed algorithm, including sample efficiency, optimality and generalization.
Specifically, we focus on the following questions: 1) {\bf Performance}: whether the proposed algorithm can outperform previous representative methods in terms of optimality and sample efficiency; 2) {\bf Ablation study}: what is the influence of different components of the proposed algorithm on the learning performance; 3) {\bf Long horizon tasks}: whether the proposed algorithm can train the multi-task agent to better solve long-horizon unseen tasks; 
4) {\bf Visualization}: what the learned Q function looks like for options conditioned different future subgoals. The neural architecture and hyper-parameters used in experiments are also introduced in Appendix.

\subsection{Experiment Setup}
\label{sec:exp-setup}
We conducted experiments across different environments and LTL tasks, where the tasks vary in length and difficulty. All the environments are procedurally generated, where the layout and positions of objects are randomly generated upon reset. The positions and properties of objects are unknown to the agent. As such, none of the environments adopted here can be solved by simple tabular-based methods. 

In every training episode, the agent uses corresponding option policies to satisfy a subgoal sequence $\xi$ randomly generated according to the current curriculum level. After every fixed number of training steps or episodes, the agent is evaluated on a fixed number (64) of tasks with LTL formulas randomly sampled from a large set of possible tasks. 
We also evaluate the agent on LTL tasks whose solution has longer horizons than subgoal sequences in the training, verifying the generalization of the trained agent to more difficult tasks. The environments are introduced in the following.

\begin{figure*}
    \centering
    \subfigure[Letter, DNF Task]{
        \centering
        \includegraphics[width=1.2in]{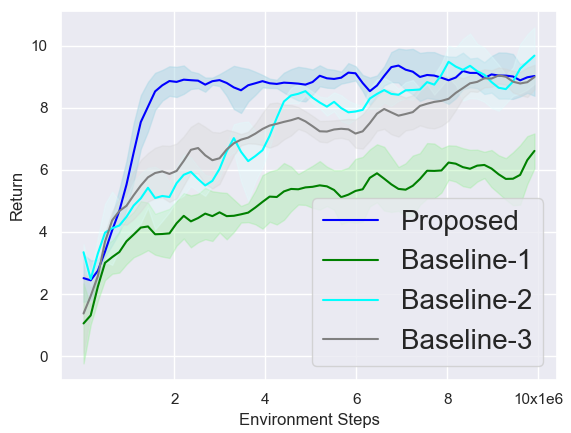}
        \label{fig:perf_letter}
    }
    \hspace{5pt}
    \subfigure[Room, DNF Task]{
        \centering
        \includegraphics[width=1.2in]{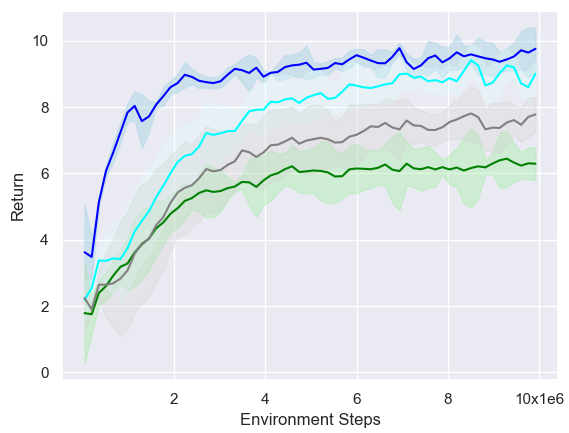}
        \label{fig:perf_room}
    }
    \hspace{5pt}
    \subfigure[Navigation, DNF Task]{
        \centering
        \includegraphics[width=1.2in]{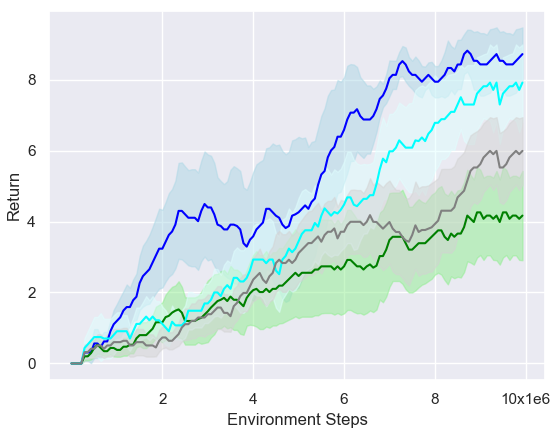}
        \label{fig:perf_navigation}
    }

    \subfigure[Letter, Rec. Task]{
        \centering
        \includegraphics[width=1.2in]{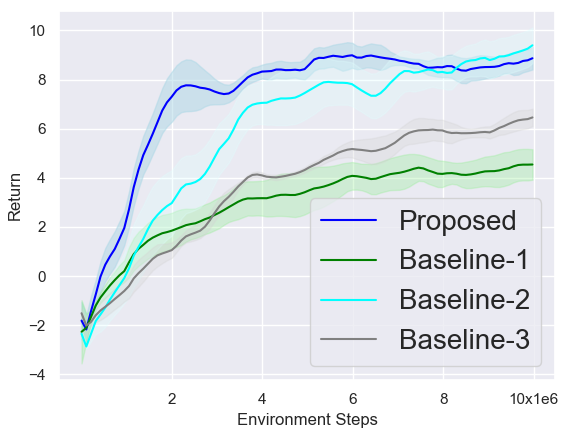}
        \label{fig:perf_letter2}
    }
    \hspace{5pt}
    \subfigure[Room, Rec. Task]{
        \centering
        \includegraphics[width=1.2in]{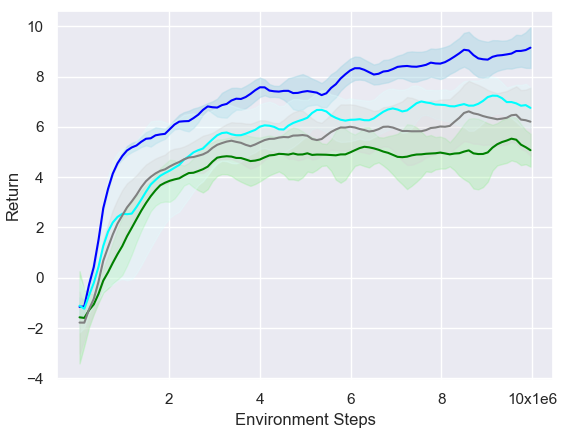}
        \label{fig:perf_room2}
    }
    \hspace{5pt}
    \subfigure[Navigation, Rec. Task]{
        \centering
        \includegraphics[width=1.2in]{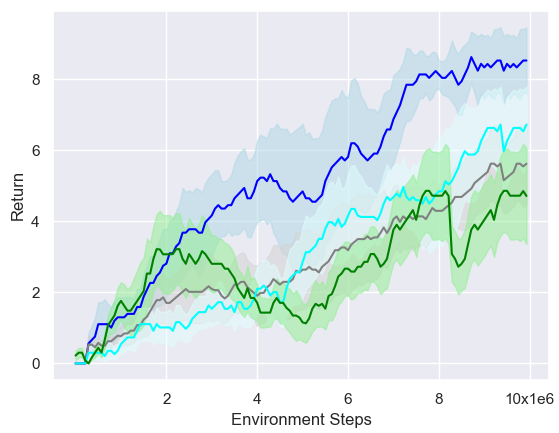}
        \label{fig:perf_navigation2}
    }
    \vspace*{-5pt}
    \caption{Performance Comparisons. "Rec." is short for recursive. The first row is for evaluating DNF tasks, and the second row is for evaluating recursive tasks. The return is defined as the sum of rewards along the trajectory.}
    \label{fig:comparison}
    \vspace*{-10pt}
\end{figure*}

\vspace{5pt}
\noindent
{\bf Letter.} This environment is a $n\times n$ grid game which is a variant of that in Figure \ref{fig:motivating}, replacing objects by letters. Out of the $n^2$ grid cells, $m$ grids are associated with $k$ (where $m>k$) unique propositions (letters). Note that some letters may appear in multiple cells, giving the option of reaching them in multiple ways. An example layout is shown in Figure \ref{fig:fsa2} with $n=7, m=10$ and $k=5$. At each step the agent can move along the cardinal directions (up, down, left and right). The agent is given the task formula and is assumed to observe the full grid (and letters) from an egocentric point of view with the image-based observation. Each task is described by an LTL formula in terms of these letters. But positions of these letters are unknown to the agent.

\vspace{5pt}
\noindent
{\bf Room.} This environment is also a grid-world game, but its observation is divided into four rooms through walls. There are 5 letters located in 8 positions, corresponding to 5 propositions randomly allocated in these rooms. An example of layout is shown in Figure \ref{fig:room}. The agent is randomly placed into one of these rooms. Each room is connected to its neighbors by corridors. Two corridors selected randomly are blocked by locks. The agent can open a lock by using a key corresponding to that specific lock (having the same color). These (green and yellow) keys are placed in positions which the agent can reach. This environment is an upgrade of MineCraft with obstacles and dependencies between objects imposed by keys and locks. The observation is also image-based here and the agent does not know the positions of objects. Every task formula is an LTL formula in terms of object's letters.

\vspace{5pt}
\noindent
{\bf Navigation.} This is a robotic environment with continuous action and state spaces. It is modified from OpenAI's Safety Gym \cite{ray2019benchmarking}. As shown in Figure \ref{fig:robot}, the environment is a 2D plane with 6 to 9 colored circles, called "navigation". Here each color represents a proposition in task specification, and some circles could share the same color. We use Safety Gym’s Point robot whose actions are for steering and forward/backward acceleration. Its observation includes the lidar information towards the circles and other sensory data (e.g., accelerometer, velocimeter). The circles and the robot are randomly positioned on the plane at the start of each episode and the robot has to visit and/or avoid certain colors in a particular manner described by the LTL specification.

\subsubsection{Tasks}
\label{sec:tasks}
We evaluate the proposed framework on two categories of tasks, and each category has millions of possible tasks. Every LTL task for testing is randomly selected, and the agent does not know any information about the task before learning starts. We define the {\it depth} of a task formula $\varphi$ as the length of shortest subgoal sequence to satisfy $\varphi$.

The first kind of task is the "DNF" task described by a disjunctive normal formula that concatenates terms by disjunctive operator $\vee$, where a term is a subgoal sequence which may have safety constraints, e.g., $\phi_{\text{DNF}}=(\Diamond(a\wedge\Diamond b)\wedge\Box\neg e)\vee\Diamond(c\wedge\Diamond d)$. Specifically, the number of terms (subgoal sequences) ranges between 3 and 6, and the number of propositions in every term is between 1 to 5. 

The second task is called "recursive" task \cite{vaezipoor2021ltl2action}, which can be formulated as $\phi_{\text{rec}}=\phi_{\text{rec}}\wedge\phi'|\phi'$ and $\phi'=\neg s\cup(g\wedge\Diamond\phi')|\neg s\cup g$. Here, $s$ and $g$ are propositions denoting two different subgoals. The notation $|$ denotes alternative. When generating a task formula, two sub-formulae around $|$ are uniformly selected. The depth of recursion is randomly selected between 3 and 5. An example of "recursive" task is $(\neg a\cup(b\wedge\Diamond(\neg c\cup d)))\wedge(\neg e\cup(f\wedge\Diamond(\neg g\cup h)))$, and the shortest subgoal sequence for satisfying this task is $b\to d$ or $f\to h$, each having depth of 2.

\subsubsection{Baselines}
\label{sec:baselines}
The proposed algorithm is compared with three baselines. The model architecture and hyper-parameters of the proposed method are introduced in Appendix \ref{sec:arch} and \ref{sec:hyper}. The first baseline ({\it Baseline-1}) is based on the conventional option framework, where a reasoning technique is used to tell the agent which proposition to achieve next and every option is trained to achieve that proposition as subgoal. This idea was widely used by previous works on multi-task RL \cite{andreas2017modular,sohn2018hierarchical,sun2019program,leon2020systematic,araki2021logical}. The agent's model here is same as that in our method, except that the options are myopic and do not consider future subgoals. In order to make comparisons to be fair, in Baseline-1 the RL algorithms for training the agent and hyper-parameters are same as the proposed method, where off-policy training, HER and formula decomposition are all adopted. However, in Baseline-1, since options are not conditioned on future subgoals and their training does not need future rewards, the multi-step value function is not needed and not used. 

\begin{figure*}
    \centering
    \subfigure[Letter, DNF Task]{
        \centering
        \includegraphics[width=1.22in]{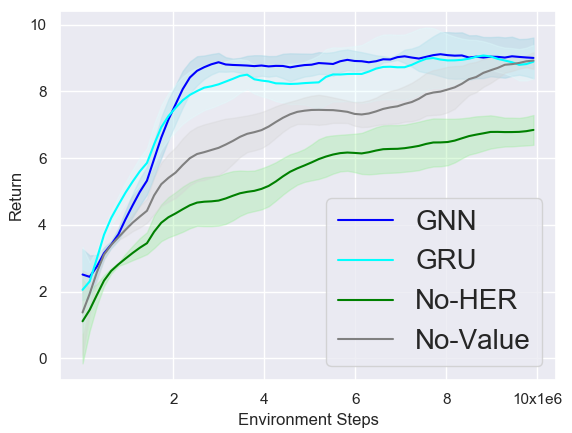}
        \label{fig:abl_letter_1}
    }
    \subfigure[Letter, Rec. Task]{
        \centering
        \includegraphics[width=1.2in]{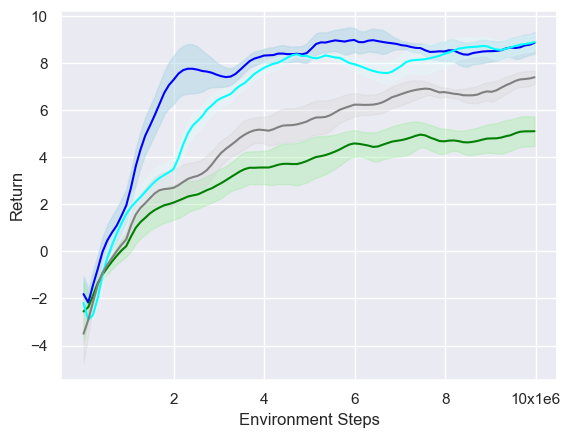}
        \label{fig:abl_letter_2}
    }
    \subfigure[Room, DNF Task]{
        \centering
        \includegraphics[width=1.2in]{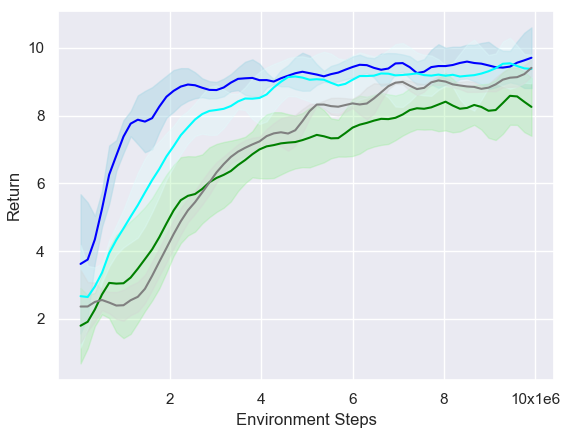}
        \label{fig:abl_room_1}
    }
    \subfigure[Room, Rec. Task]{
        \centering
        \includegraphics[width=1.2in]{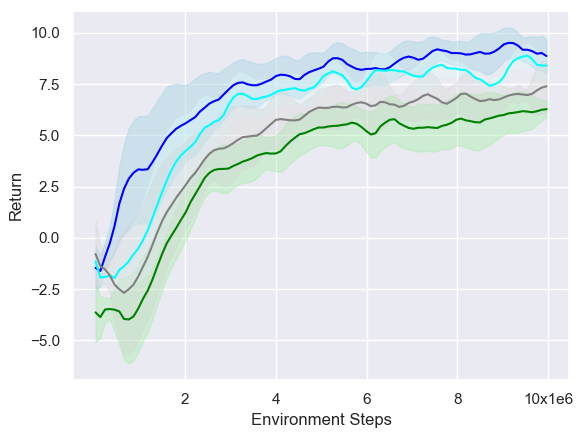}
        \label{fig:abl_room_2}
    }
    \vspace*{-5pt}
    \caption{Ablation study. "No-HER" refers to the proposed method without using HER. "No-value" refers to the proposed method without using the multi-step value function.}
    \label{fig:abl}
    \vspace*{-10pt}
\end{figure*}

The second baseline ({\it Baseline-2}) is modified from \cite{vaezipoor2021ltl2action}, where the task formula is processed by a graph convolutional network (GCN) \cite{kipf2016semi} and progresses over time. The architecture of GCN here is the same as that in \cite{vaezipoor2021ltl2action} with $T=8$ message passing steps and 32-dimensional node embedding. Other parts of agent's model are the same as the proposed method. The third baseline ({\it Baseline-3}) is based on the method in \cite{kuo2020encoding}. This approach trains an agent that considers the whole task formula as an extra input and uses GRU \cite{chung2014empirical} to learn an embedding of the LTL formula which does not progress over time. The learned task embedding has the size of 32 which is same as the size of embedding of future subgoals in our method. Other parts of agent's model are the same as the proposed method. 

In original papers of Baseline-2 and 3 \cite{kuo2020encoding,vaezipoor2021ltl2action}, the agent is trained by on-policy PPO algorithms, which are not as sample-efficient as their off-policy counterparts, and hence do not fit for the comparison with our method. As such, the agents in Baseline-2 and 3 are trained by off-policy Q learning \cite{mnih2015human} or SAC \cite{haarnoja2018soft} approach which use the same hyper-parameters as the proposed method. Since the agent takes the original LTL formula as its input directly, formula decomposition and HER cannot be used in Baseline-2 or 3. Since their original implementations are not option-based, the multi-step value function $V^{\phi}$ is not used either.

\subsection{Results}
\label{sec:results}
In this section, we present the comparison results of the proposed method with baselines. The overall performance comparisons in terms of average return for satisfying LTL tasks are first presented. Then, we present the ablation studies to investigate the effects of different components of the proposed method. 
The results for {\it long-horizon tasks} and {\it visualization} are introduced in Appendix \ref{sec:propagation} and \ref{sec:visualization}, respectively.

\subsubsection{Performance}
\label{sec:performance}
In Figure \ref{fig:comparison}, the proposed method is compared with three baselines introduced in Section \ref{sec:baselines}. We can see that although Baseline-1 can learn fast in the early stage, its overall performance is the worst. The optimality in Baseline-1 degrades because the resulting options myopically focus on the next subgoal only, without looking ahead. It shows the importance of the dependence of options on future subgoals. In addition, the proposed method can learn much faster than Baseline-2 and 3, showing that leveraging reusable skills via options can achieve better sample efficiency. The agents in Baseline-2 and 3, which are conditioned on the LTL formula directly, need a lot of environment samples to understand temporal operators and find out the optimal path in the formula to finish the task.

\subsubsection{Ablation Study}
\label{sec:ablation}

The ablation study is first to compare using GNN or GRU in option critics $Q^{\theta}_p(\cdot,\cdot;\xi)$ and value function $V^{\phi}(\cdot;\xi)$ to learn the embedding of the sequence $\xi$ of future subgoals. Specifically, the nodes of GNN represent subgoals and every subgoal is connected to its successor by a directed edge. The embedding of sequence $\xi$ is learned by GCN with multi-step message passing ($T=8$). In addition, when the GRU is used, sequence $\xi$ with every element on-hot encoded is fed into GRU and the embedding can be obtained at the output of GRU. More details of agent's model are in Appendix. In Figure \ref{fig:abl}, we can see that the GRU performs slightly worse than GNN.

In addition, we also study the effects of multi-step value function $V^{\phi}$ and HER by comparing "No-value" and "No-HER" with the proposed method in Figure \ref{fig:abl}. We can see that when $V^{\phi}$ or HER is not used, the learning performance can degrade significantly. Furthermore, the performance degradation of "No-HER" is larger, showing that the performance improvement from HER is more than that from $V^{\phi}$. This is because the trajectories relabeled by HER are used for training both value function $V^{\phi}$ and Q function $Q^{\theta}$.

\section{Conclusion}
In this work, we propose a novel framework for generalizing LTL instructions by options dependent on the future subgoal sequence. Moreover, to facilitate the reward propagation of satisfying future subgoals, we propose to learn a multi-step value function updated by Monte Carlo estimates of discounted return. With comprehensive experiments, the proposed method is confirmed to have significant advantages over previous methods in terms of optimality and sample efficiency.

\bibliography{main}

\newpage
\onecolumn

\section{Appendix}
\label{sec:appendix}

\subsection{Practical Implementation}
\label{sec:practical2}
Here we introduce more details on the practical implementation. The training curriculum and HER are used in the training Algorithm \ref{alg:training}. The LTL decomposition, execution and safety shield are used in the testing Algorithm \ref{alg:testing}. 

\vspace{5pt}
\noindent
{\bf Training Curriculum.} During training, the agent is trained to satisfy a randomly generated subgoal sequence $\xi$ with maximal environment return.
Denote the maximum length of $\xi$ as $K$. The training curriculum consists of $K$ levels. As such, in the $k$-th level ($k=1,\ldots,K$), the length of subgoal sequence $\xi$ is set to be $k$. And when his average success rate in $k$-th level is above a threshold (e.g., $80\%$), the agent will proceed to $(k+1)$-th level. Therefore, the difficulty of tasks increases gradually as the agent proceeds to higher levels. For any subgoal sequence $\xi$, the agent applies options to satisfy subgoals in $\xi$ one-by-one with conditions of future subgoals. The details are introduced in Algorithm \ref{alg:training}.

We also adopt an adversarial scheme for selecting training tasks which can improve the learning efficiency in empirical experiments. In the $k$-th level, at the beginning of each episode with initial state $s_0$, multiple subgoal sequences with same length are randomly generated, i.e., $\{\xi_i\}_{i=1}^{N_T}$, and the $j$-th sequence with lowest value is selected as the training task for the agent, i.e., $j=\arg\min_{i=1,\ldots,N_T}V^{\phi}(s_0;\xi_i)$. It means that a difficult task in current level is selected to train the agent, always pushing forward the capability of the agent.

\vspace{5pt}
\noindent
{\bf LTL Task Decomposition. } During testing, the agent is tested to satisfy a random unseen task which is described by an LTL formula $\varphi$ in terms of propositions in $\mathcal{P}$. The agent first decomposes $\varphi$ into a list of finite sequences consisting of subgoals in $\mathcal{P}$, which is based on the decomposition algorithm introduced in \cite{leon2020systematic,leon2021nutshell}. Specifically, it transforms the complex LTL formula $\varphi$ into a list $\mathcal{K}$ consisting of all the sequences $\tau_i$ of subgoals $p\in\mathcal{P}$ that satisfy $\varphi$, i.e., $\mathcal{K}=\{\xi_i\}_{i=1}^{M_{\varphi}}$ where $M_{\varphi}$ is the number of subgoal sequences that satisfy $\varphi$. Note that in a given sequence $\xi_i$ here, every symbol or proposition is not redundant and makes progress towards the satisfaction of task formula $\varphi$. The detailed decomposition algorithm is introduced in Appendix for completeness. 

\vspace{5pt}
\noindent
{\bf Execution.} After decomposing formula $\varphi$ into list $\mathcal{K}$, we first select the sequence $\xi_{i^*}$ with the largest value and apply corresponding options to execute it, i.e., $i^*=\arg\max_{i=1,\ldots,M_{\varphi}}V^{\phi}(s_0;\xi_i)$ where $s_0$ is the initial state. In every step $t$, we examine the current symbolic observation $\sigma_t$, i.e., $\sigma_t=L(s_t)$, and use $\sigma_t$ to progress the LTL formula $\varphi$, i.e., $\varphi\leftarrow\text{prog}(\sigma_t,\varphi)$. For every $\xi\in\mathcal{K}$, if $\sigma_t$ can entail the first subgoal $\xi[0]$, then $\xi[0]$ is removed from $\xi$, i.e., $\xi\leftarrow\xi[1:]$. Then, whenever the optimal sequence $\xi_{i^*}$ is selected based on $V^{\phi}(s_t;\cdot)$, we apply the corresponding option policy $\pi_{\xi_{i^*}[0]}^{\xi_{i^*}[1:]}$ to execute.

\vspace{5pt}
\noindent
{\bf Safety Shield.} 
Our framework can be easily used to satisfy safety constraints specified by the task formula. The safety constraint implies that the agent needs to avoid any propositions which would falsify the task formula whenever it is applying the option policy.
Whenever $\varphi$ is progressed, a set of unsafe propositions $\mathcal{U}$ can be constructed as $\mathcal{U}:=\{q|q\in\mathcal{P}, \text{prog}(q,\varphi)=\text{false}\}$. If the selected action $a_t$ can lead the agent to any proposition in $\mathcal{U}$, i.e., $\exists q\in\mathcal{U}$ and $Q^{\theta}_q(s_t,a_t;\varnothing)>\kappa$, the agent has to sample and select a new action until the action becomes safe. Here $\kappa$ is the threshold of being close to any proposition.

\subsection{Training and Testing Algorithms}
\label{sec:algos}
We summarize the detailed operations in training and testing the multi-task agent in Algorithm \ref{alg:training} and \ref{alg:testing}, respectively.

\begin{algorithm}[ht]
\caption{Training Multi-task Agent for Following LTL Instructions}
\label{alg:training}
\begin{algorithmic}[1]
\State {Environment MDP $\mathcal{M}_e$; labeling function $L$; positive reward for task completion $R_F$; The set of propositions $\mathcal{P}$; value function $V^{\phi}(s;\xi)$; Q function of option policy $Q^{\theta}_p(s,a;\xi)$ for $\forall p\in\mathcal{P}$; replay buffer $\mathcal{B}$; trajectory buffer $\mathcal{B}_t$ episodic buffer $\mathcal{E}$; maximum length of subgoal sequence $K$; performance threshold $\zeta$ of upgrading to next level}
\State Initialize parameters $\theta$ and $\phi$;
\State Initialize $\mathcal{B}\leftarrow []$;
\State \% levels from 1 to $K$;
\For{$k=1,\ldots,K$ } 
\While{the average success rate is below $\zeta$}
\State{Initialize $\mathcal{E}\leftarrow []$;}
\State Reset environment $s\leftarrow s_0$;
\State{Randomly generate $N_S$ subgoal sequences, and select $\xi$ with lowest value on $V^{\phi}$;}
\For{$l=1,\ldots,\text{len}(\xi)$} 
{\it \# $\text{len}(\xi)$ denotes the length of $\xi$}
\State $\tilde{s}_0\leftarrow s$;
\For{$t=0,\ldots,T_S-1$}
\State Apply option policy $\pi^{\xi[1:]}_{\xi[0]}$ into the environment $\mathcal{M}_e$;
\State Obtain reward $r_t$ and next state $\tilde{s}_{t+1}$;
\State {Store experience tuple $(\tilde{s}_t,a_t,r_t,\tilde{s}_{t+1},\xi[0],\xi[1:])$ into $\mathcal{E}$ and $\mathcal{B}$;}
\If {$L(\tilde{s}_{t+1})\models\xi[0]$}
\State Set $s\leftarrow\tilde{s}_{t+1}$ and $\xi\leftarrow\xi[1:]$;
\State Go to 10; 
\EndIf
\State Sample a minibatch $\mathcal{B}_M$ from $\mathcal{B}$ and update Q function according to \eqref{q-train};
\State Sample trajectories from $\mathcal{B}_t$ and update $V^{\phi}$ according to \eqref{v-train};
\EndFor
\State Break; {\it \# the trajectory $\mathcal{E}$ is unsuccessful and needs to be relabeled}
\EndFor
\If {$\mathcal{E}$ is unsuccessful} {\it \# relabel unsuccessful trajectory}
\State Randomly select subgoal sequence $\xi'$ satisfied by $\mathcal{E}$;
\State Relabel the subgoal and condition of every tuple in $\mathcal{E}$ based on $\xi'$;
\EndIf
\State Store transitions of $\mathcal{E}$ into $\mathcal{B}$;
\State Store $\mathcal{E}$ into $\mathcal{B}_t$;
\EndWhile
\EndFor
\end{algorithmic}
\end{algorithm}

\begin{algorithm}[ht]
\caption{Testing Multi-task Agent for Following LTL Instructions}
\label{alg:testing}
\begin{algorithmic}[1]
\State {The environment $\mathcal{M}_e$; labeling function $L$; the set of propositions $\mathcal{P}$; progression function $\text{prog}(\cdot,\cdot)$ introduced in Section \ref{sec:rl_ltl}; value function $V^{\phi}$ and critics of options $Q^{\theta}_p$ for $\forall p\in\mathcal{P}$ trained by Algorithm \ref{alg:training}; the threshold of closeness $\kappa$; the test LTL formula $\varphi$;}
\State Reset environment and obtain the initial state $s_0$;
\State Given $\varphi$, decompose it into a set $\mathcal{K}=\{\xi_i\}_{i=1}^{M_{\varphi}}$ of accepting subgoal sequences by using Algorithm \ref{alg:decomposing};
\State Given $\varphi$, obtain the set of unsafe propositions $\mathcal{U}$;
\State Select $\xi^*$ with largest value such that $\xi^*=\arg\max_{\xi\in\mathcal{K}}V^{\phi}(s_0;\xi)$;
\State set $t\leftarrow 0$;
\While{every sequence $\xi\in\mathcal{K}$ is not empty}
\State Sample action $a_t$ from the option policy $\pi_{\xi^*[0]}^{\xi^*[1:]}(\cdot|s_t)$ until $\forall q\in\mathcal{U}, Q_q^{\theta}(s_t,a_t;\varnothing)<\kappa$
\State Obtain next state $s_{t+1}$;
\If{$L(s_{t+1})\models\xi^*[0]$}
\State Progress the formula $\varphi\leftarrow\text{prog}(L(s_{t+1}), \varphi)$
\State Update the set $\mathcal{U}\leftarrow\{q|q\in\mathcal{P}, \text{prog}(q, \varphi)=\text{false}\}$;
\State $\forall\xi\in\mathcal{K}$, if $L(s_{t+1})\models\xi[0]$, then $\xi.\text{pop}(\xi[0])$
\State Select again $\xi^*=\arg\max_{\xi\in\mathcal{K}}V^{\phi}(s_{t+1};\xi)$;
\EndIf
\State $t\leftarrow t+1$
\EndWhile
\end{algorithmic}
\end{algorithm}

\begin{algorithm}[ht]
\caption{Decomposition of LTL formula into a list of subgoal sequences}
\label{alg:decomposing}
\begin{algorithmic}[1]
\State LTL formula $\varphi$;  the set of propositions $\mathcal{P}$;
\State Initialize $\mathcal{T}\leftarrow\{([], \varphi)\}_{i=1}^{|\mathcal{P}|}$;
\State Initialize $\mathcal{K}\leftarrow \{\}$;
\While{true}
\State expanded$\leftarrow$false;
\For{$(\xi, \varphi)$ in $\mathcal{T}$}
\State \# expand the tuple $(\xi,\varphi)$;
\For{$p\in\mathcal{P}$}
\If{$\text{prog}(p,\varphi)\neq\varphi$ and $\text{prog}(p,\varphi)\neq$false}
\State expanded$\leftarrow$true;
\State $\xi'\leftarrow\xi$.append($p$);
\State $\varphi'\leftarrow\text{prog}(p, \varphi)$;
\State $\mathcal{T}$.append($(\xi', \varphi')$);
\If{$\varphi'==$true}
\State $\mathcal{K}$.append($\xi'$);
\EndIf
\EndIf
\EndFor
\State $\mathcal{T}$.remove($(\xi, \varphi)$)
\EndFor
\If{expanded $==$ false}
\State Break; \# if none of tuples in $\mathcal{T}$ is expandable, stop the algorithm
\EndIf
\EndWhile
\State {\bf Return} $\mathcal{K}$
\end{algorithmic}
\end{algorithm}

\subsection{LTL Decomposition Algorithm}
For completeness, we include the algorithm for decomposing an LTL formula $\varphi$ into the list $\mathcal{K}$ of subgoal sequences which satisfy $\varphi$, which is modified from \cite{leon2021nutshell}. It uses breadth-first-search to find all the satisfying subgoal sequences. We maintain a search tree $\mathcal{T}$, each element of which is a tuple consisting of a subgoal sequence $\xi$ and formula $\varphi$ to be satisfied. The details are presented in Algorithm \ref{alg:decomposing}.

\subsection{Experiment Results: Long Horizon Tasks}
\label{sec:propagation}
In order to verify the effectiveness of the reward propagation, we evaluate the performance of the trained RL agent in tasks with long time horizon. We focus on the letter domain where the map size and the depths of the LTL formula are changed for comparison. The depth of a formula $\varphi$ is the length of optimal subgoal sequence to satisfy $\varphi$. Baseline-1 only learns independent option for each subgoal and does not consider reward propagation. Baseline-3 uses recurrent GNN to process the LTL formula not progressed, so its performance on LTL tasks with long horizon is much worse than Baseline-2. Therefore, we do not consider Baseline-1 and Baseline-3 for comparison here. In every experiment, there are 8 unique letters on the map and every letter appears twice. 

\begin{figure*}
\centering
\subfigure[$n=9, d=10$]{
    \centering
    \includegraphics[width=1.2in]{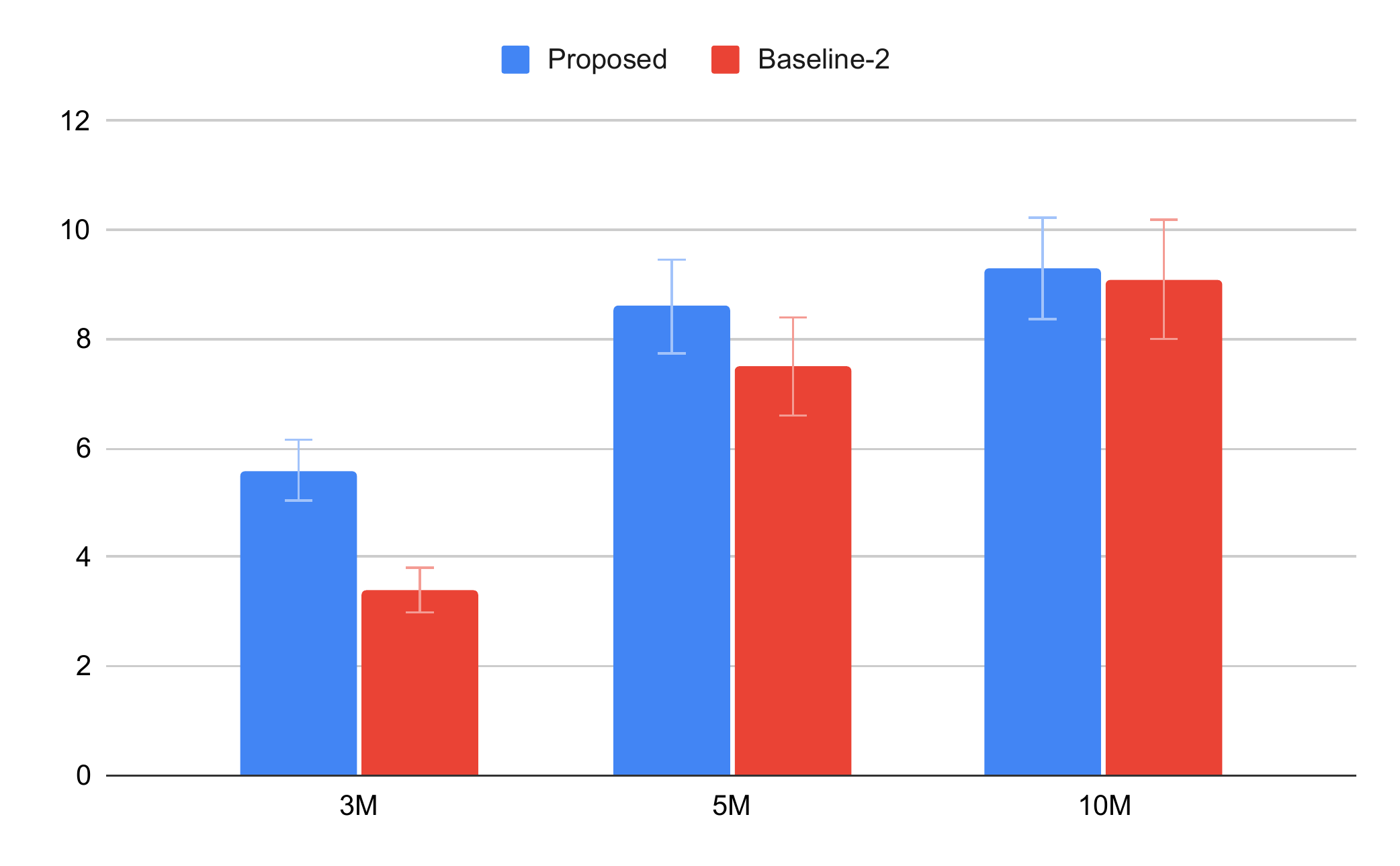}
}
\subfigure[$n=9, d=12$]{
    \centering
    \includegraphics[width=1.2in]{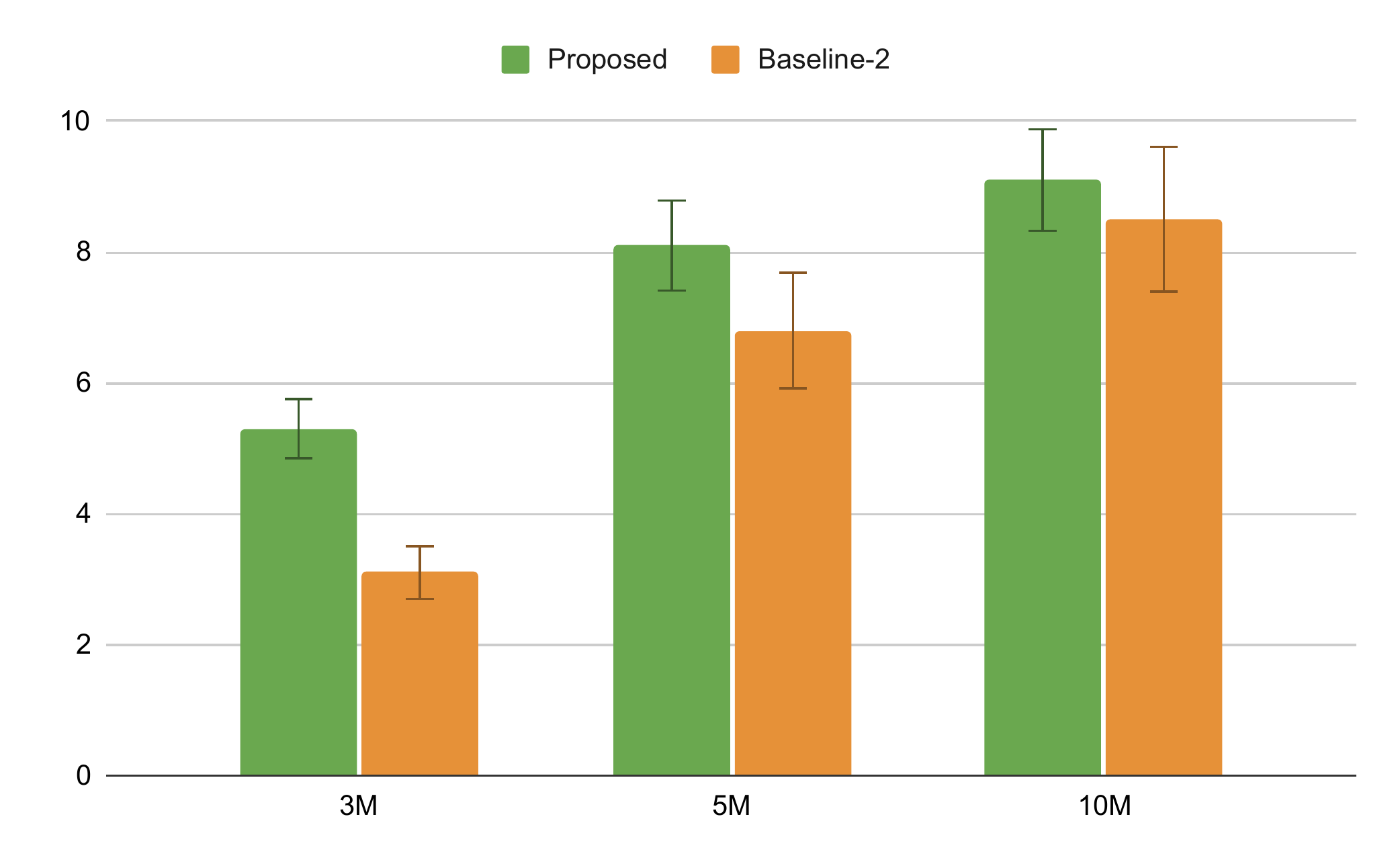}
}
\subfigure[$n=9, d=15$]{
    \centering
    \includegraphics[width=1.2in]{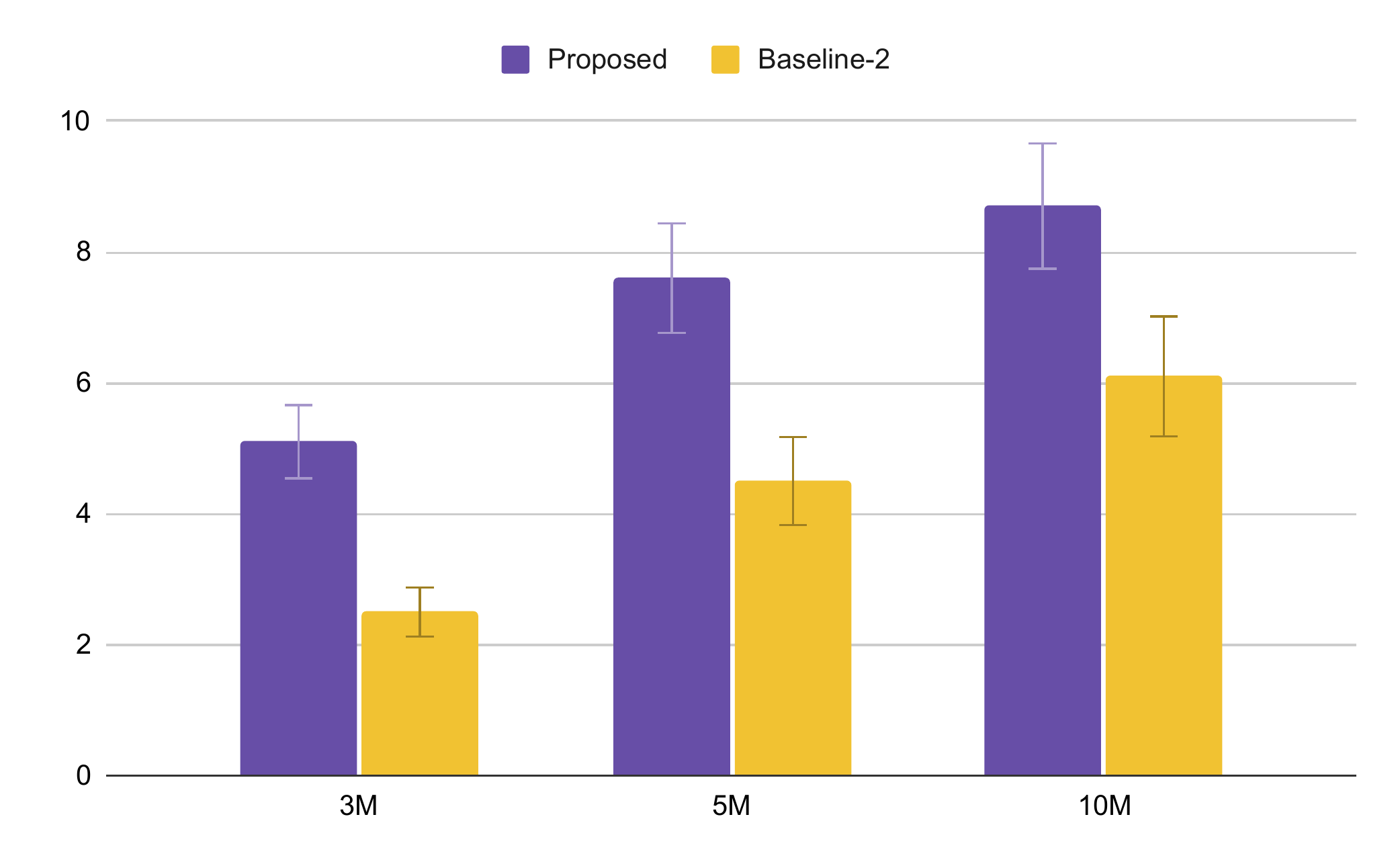}
}

\subfigure[$n=11, d=10$]{
    \centering
    \includegraphics[width=1.2in]{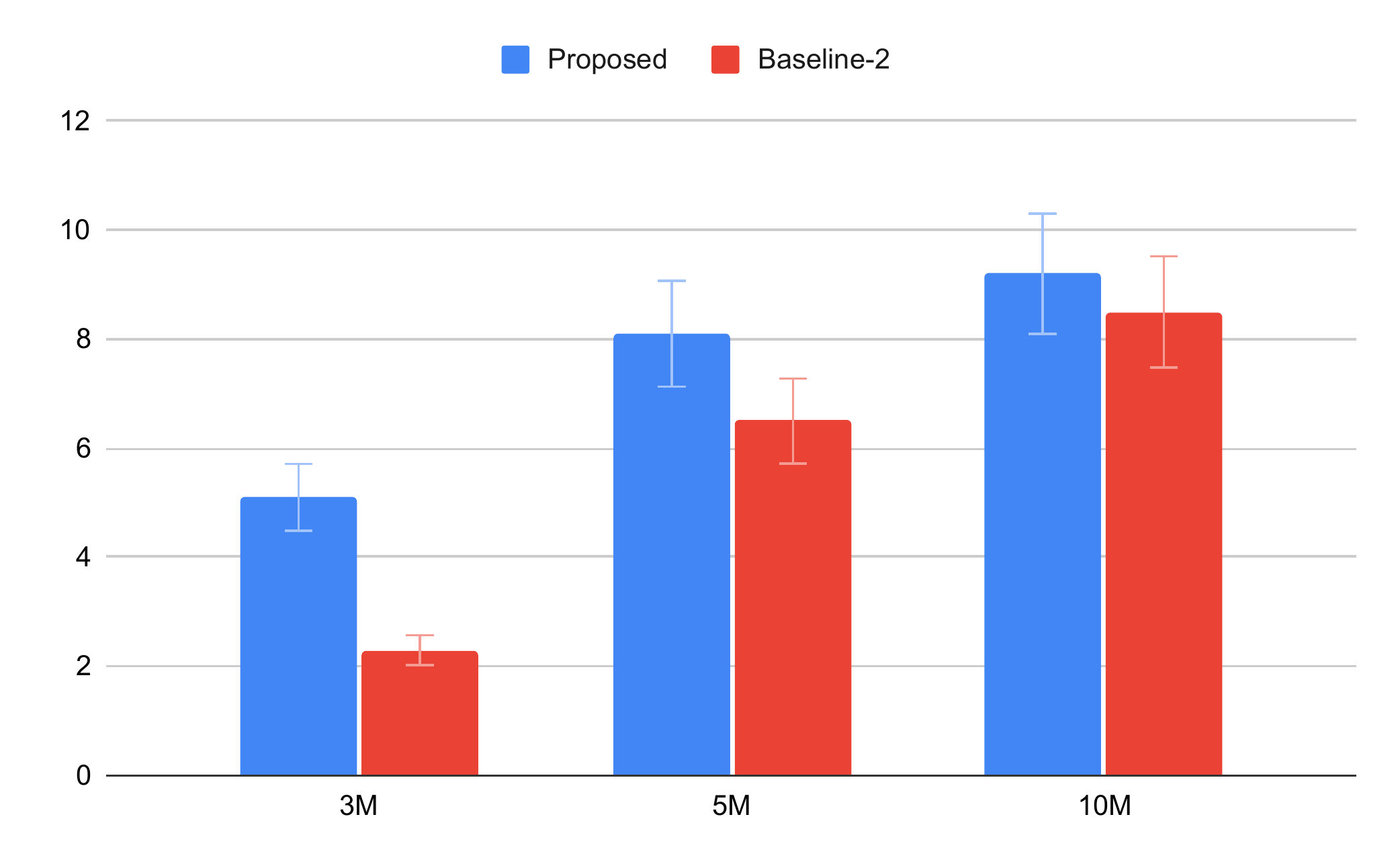}
}
\subfigure[$n=11, d=12$]{
    \centering
    \includegraphics[width=1.2in]{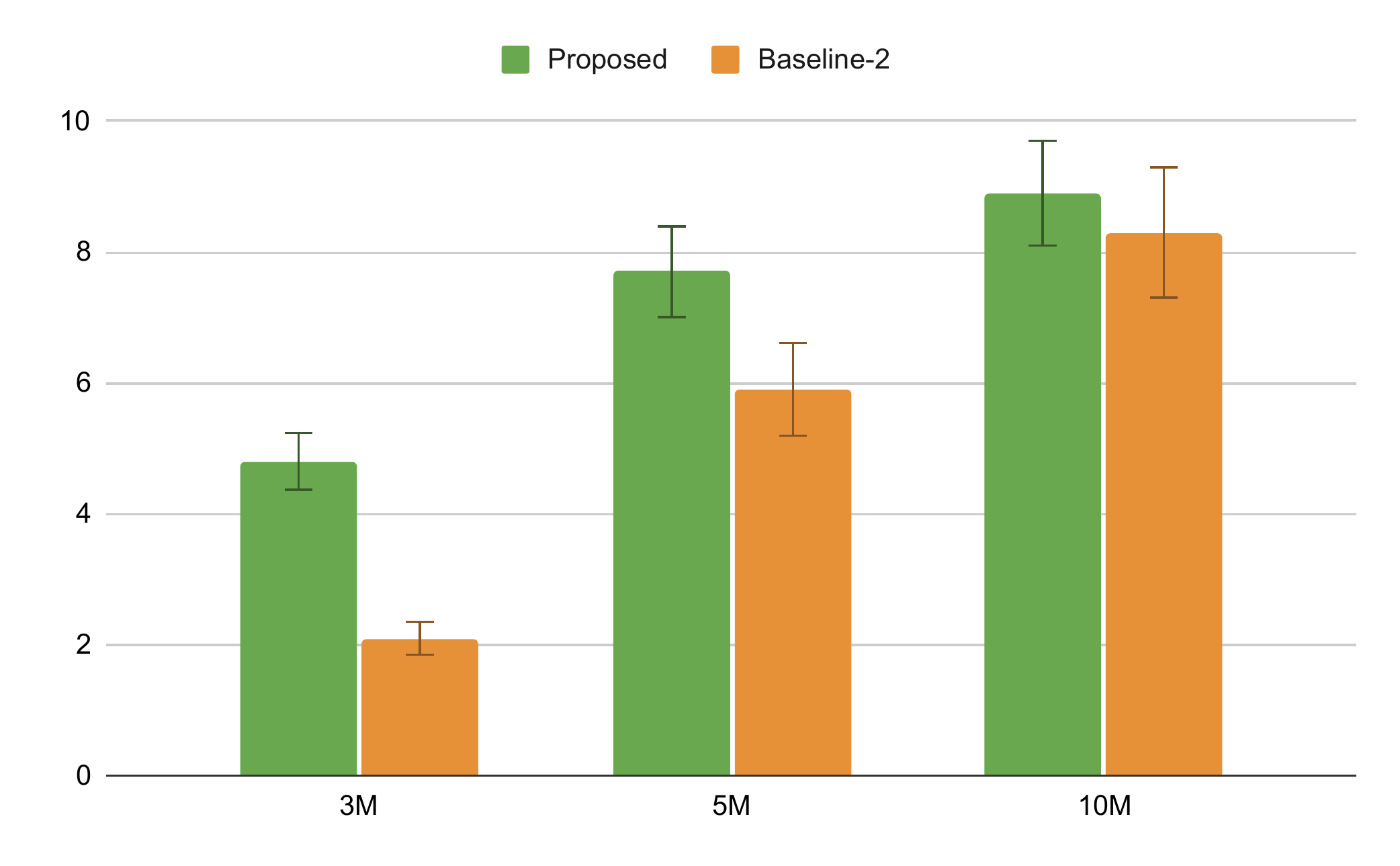}
}
\subfigure[$n=11, d=15$]{
    \centering
    \includegraphics[width=1.2in]{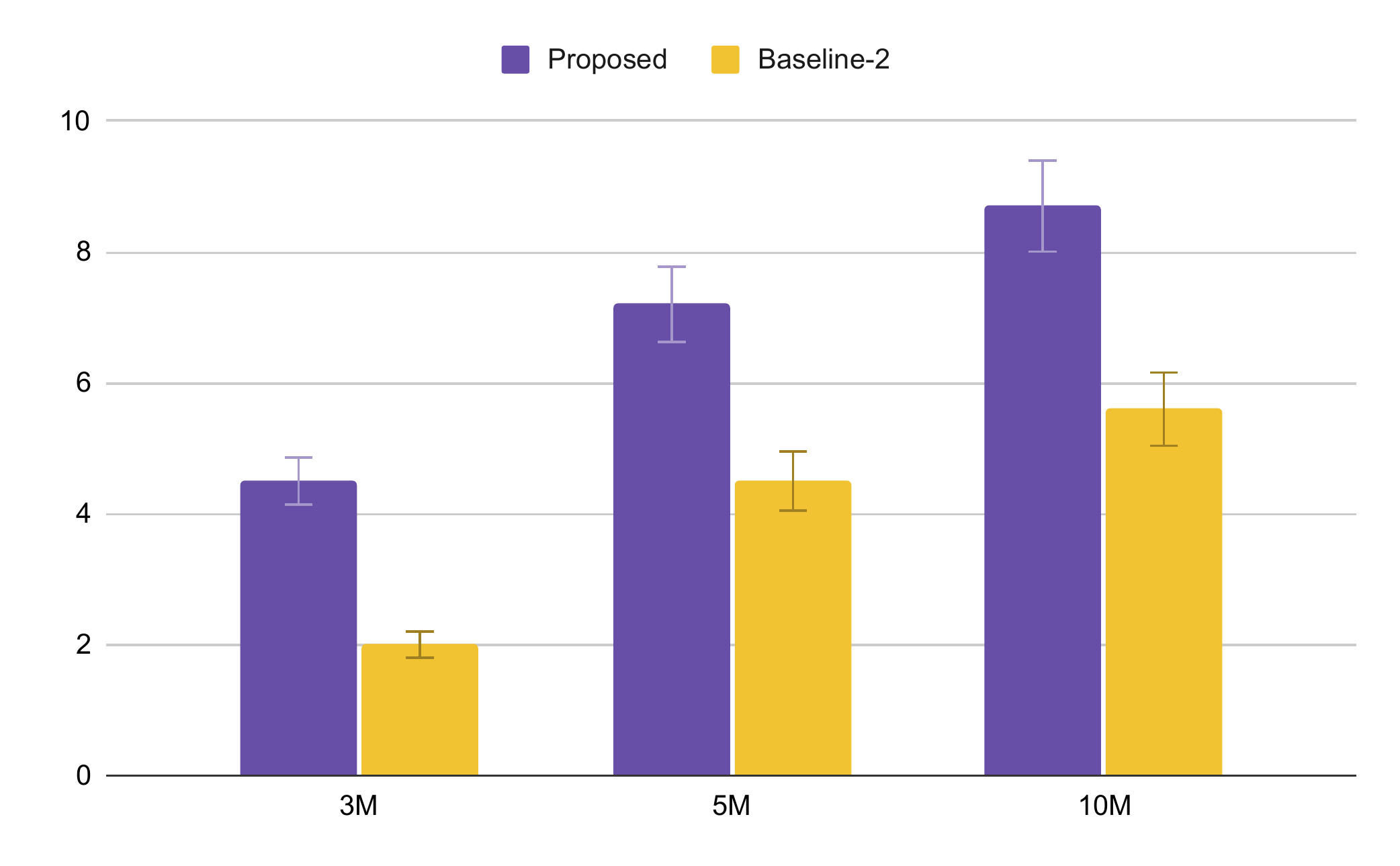}
}

\subfigure[$n=15, d=10$]{
    \centering
    \includegraphics[width=1.2in]{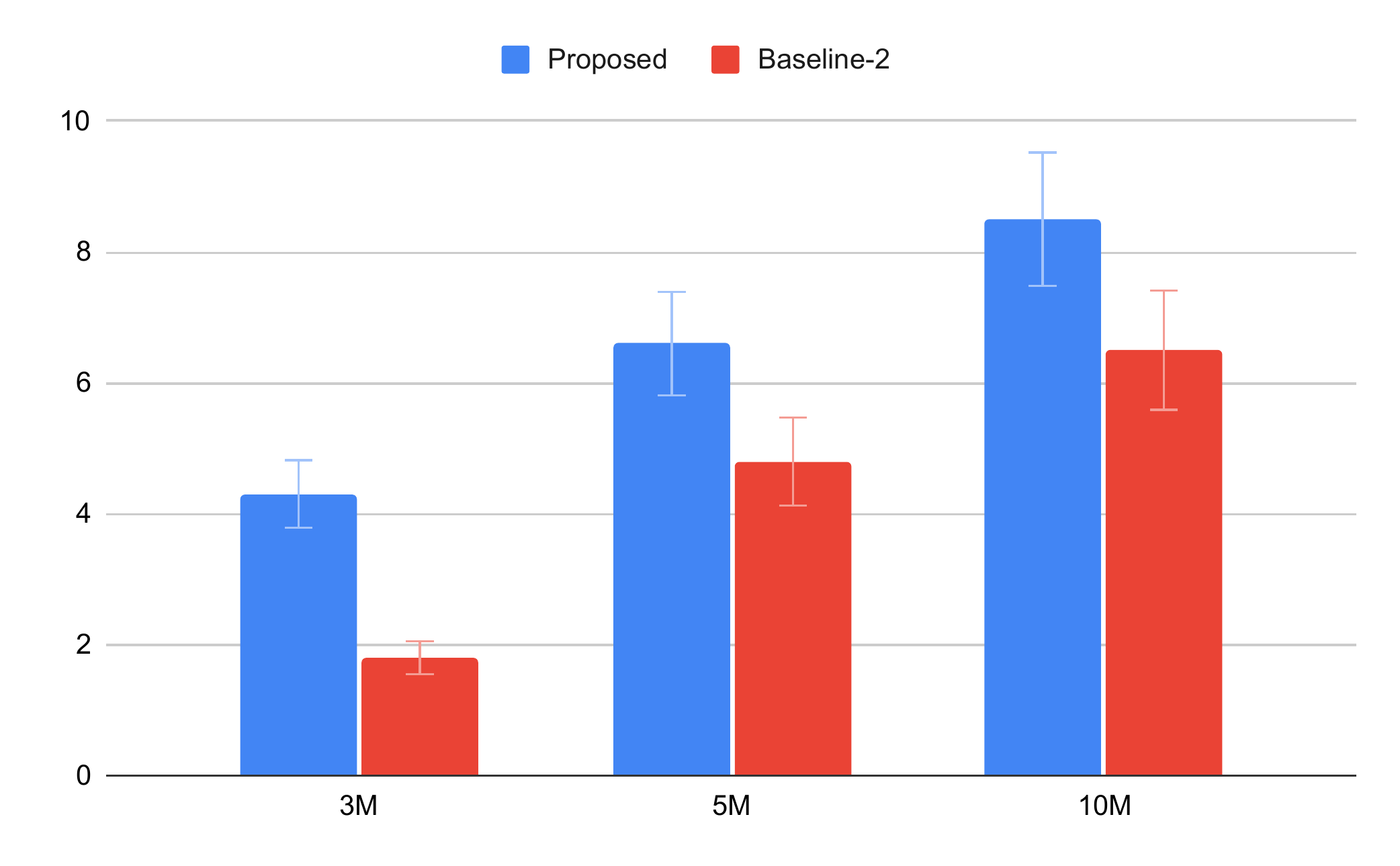}
}
\subfigure[$n=15, d=12$]{
    \centering
    \includegraphics[width=1.2in]{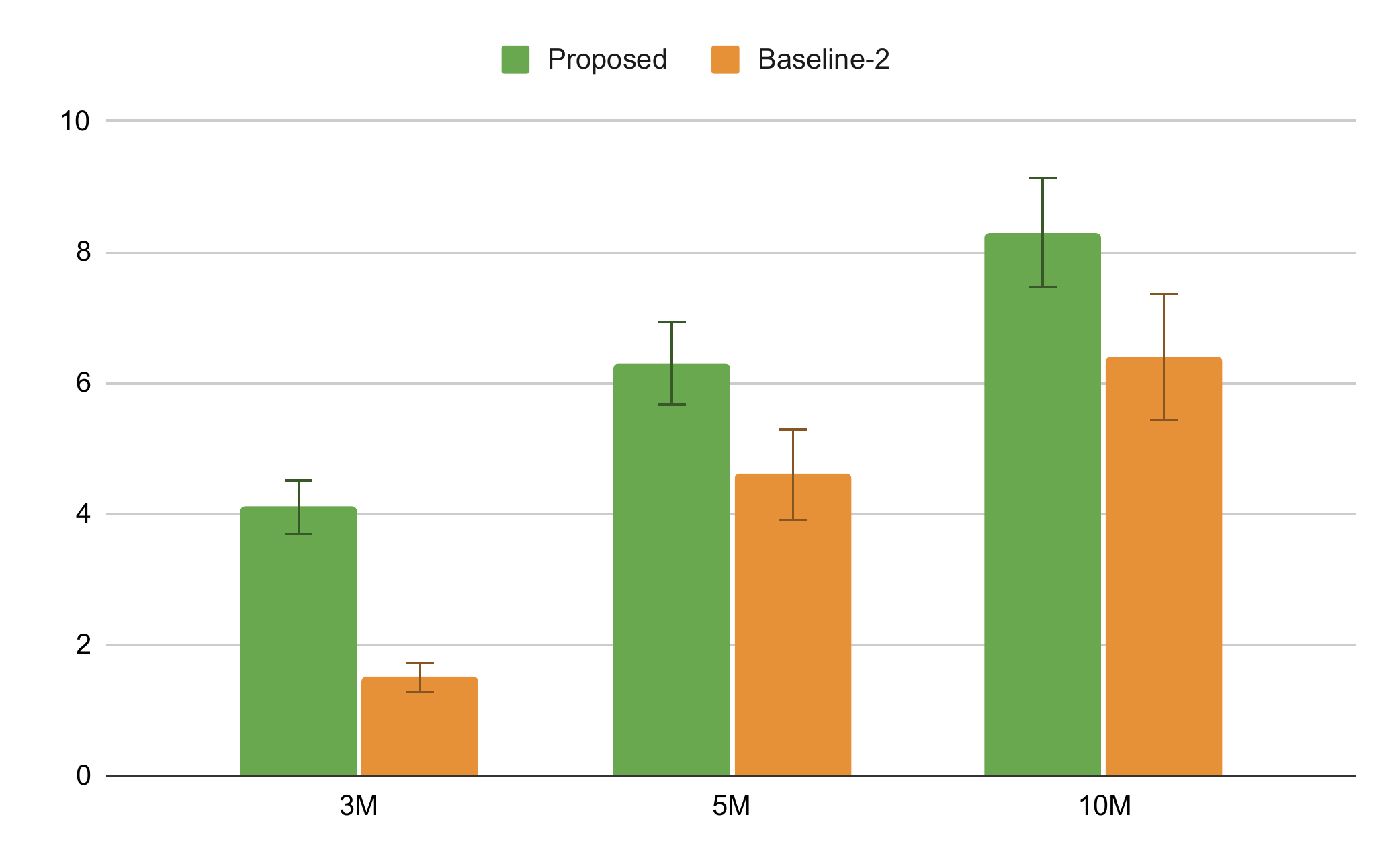}
}
\subfigure[$n=15, d=15$]{
    \centering
    \includegraphics[width=1.2in]{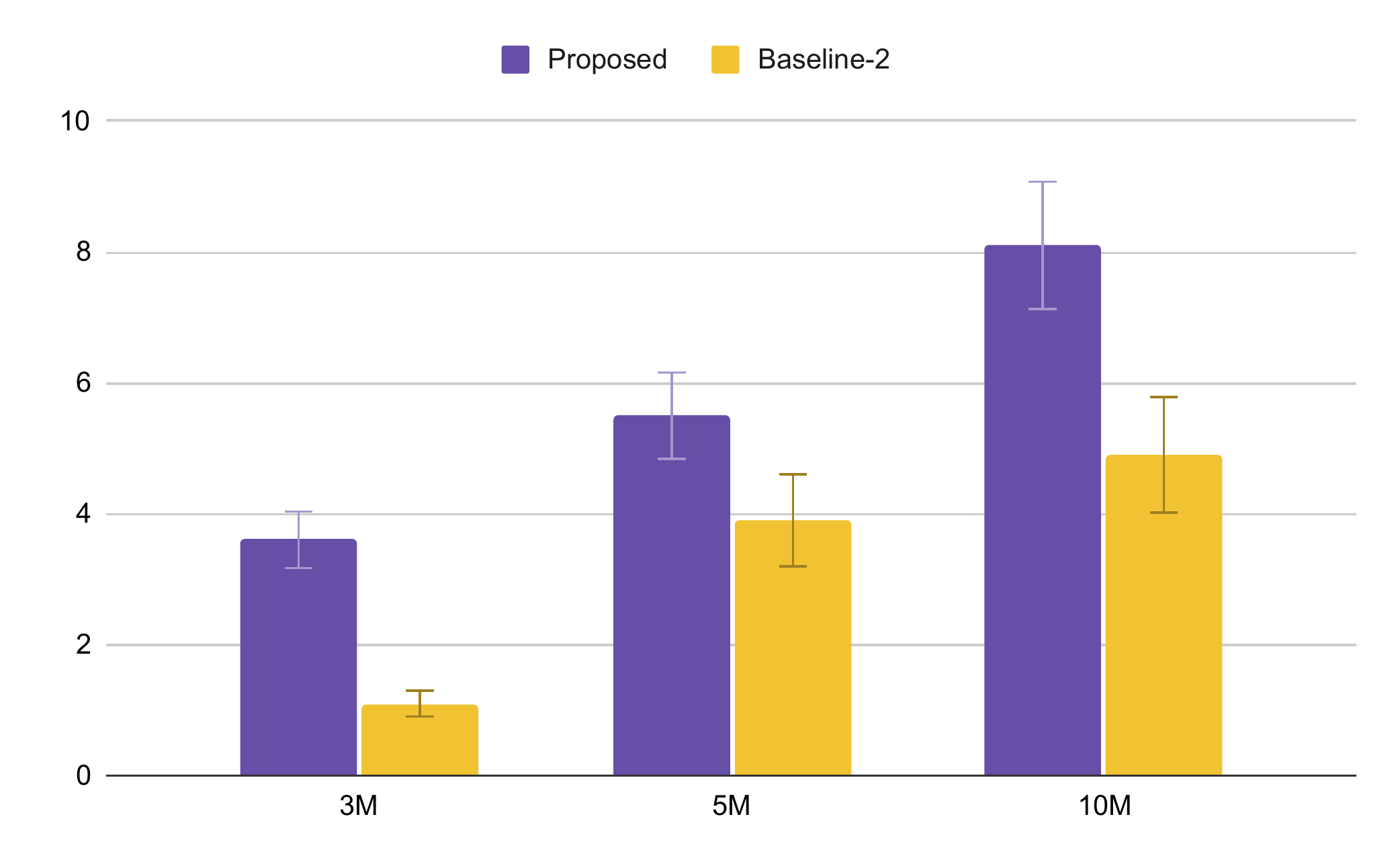}
}
\caption{Performance comparison for long-horizon tasks in letter domain. The map size is $n\times n$ and the task formula has depth of $d$. The evaluation takes place at the environment steps of $\{3, 5, 10\}\times 10^6$, during agent's training. The y-axis is the average sum of rewards received in the trajectory.}
\label{fig:reward}
\end{figure*}

The comparison results are shown in Figure \ref{fig:reward}. Since the evaluation results of long-horizon tasks have large variances, we only show the results as charts here. The LTL formula for evaluation is a DNF task consisting of 3 conjunctions with depth of $d$, where every letter is randomly generated without repetition. Every task here has longer horizon than that in Figure \ref{fig:comparison}. Every result in Figure \ref{fig:reward} is the average of 10 formulas, and the variance is obtained from 5 seeds. We can see that the proposed method can significantly outperform the Baseline-2, and the advantage increases with map size and formula depth, showing that the proposed method can solve the long-horizon tasks well and the effect of reward propagation is significant. 

\subsection{Experiment Results: Visualization}
\label{sec:visualization}

\begin{figure*}
    \centering
    \subfigure[Letter]{
        \centering
        \fontsize{6pt}{10pt}\selectfont
        \def\svgwidth{1.in}
        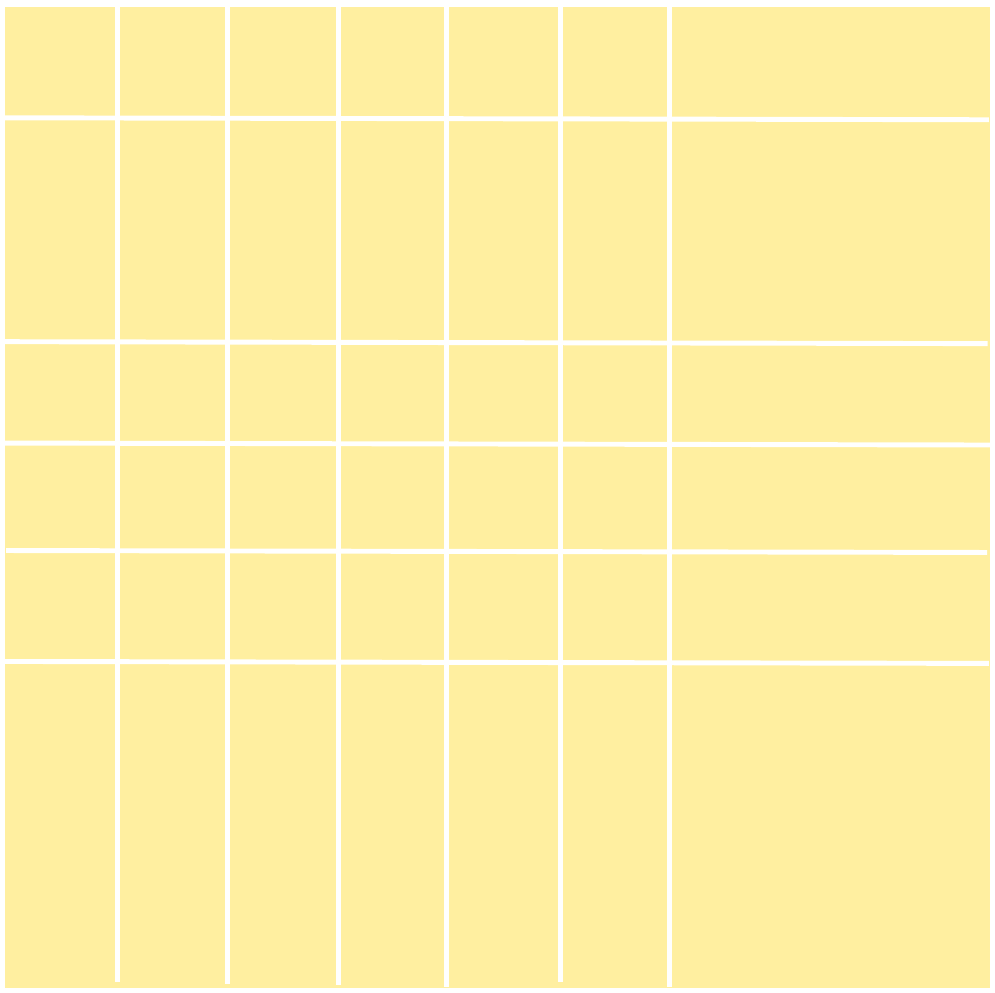
        \label{fig:vis_letter_env}
    }
    \hspace{5pt}
    \subfigure[$Q^{\theta}_a(\cdot,*;\varnothing)$]{
        \centering
        \includegraphics[width=1.in, height=1.in]{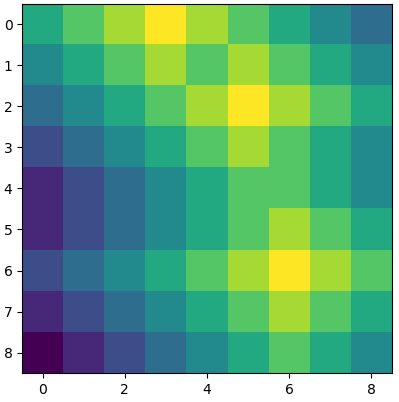}
        \label{fig:vis_letter_0}
    }
    \hspace{5pt}
    \subfigure[$Q^{\theta}_a(\cdot,*;b)$]{
        \centering
        \includegraphics[width=1.in, height=1.in]{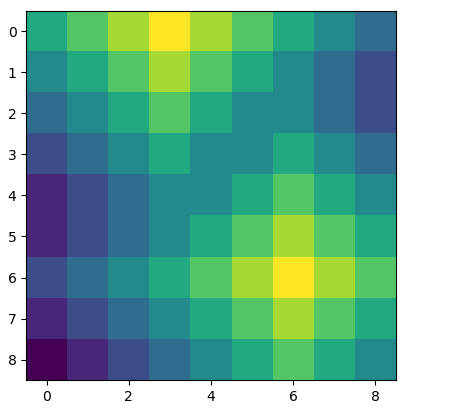}
        \label{fig:vis_letter_1}
    }
    \hspace{5pt}
    \subfigure[$Q^{\theta}_a(\cdot,*;b,c)$]{
        \centering
        \includegraphics[width=1.in, height=1.in]{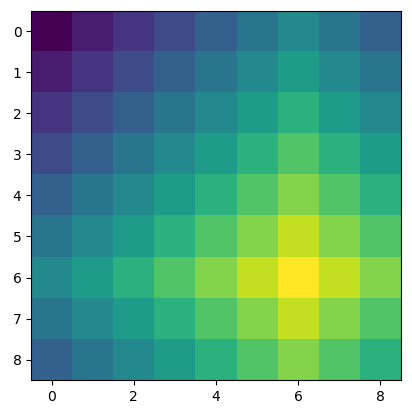}
        \label{fig:vis_letter_2}
    }
    
    \subfigure[Room]{
        \centering
        \includegraphics[width=1.in, height=1.in]{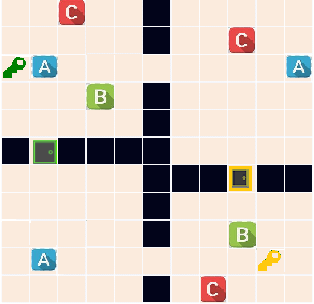}
        \label{fig:vis_room_env}
    }
    \hspace{5pt}
    \subfigure[$Q^{\theta}_C(\cdot,*;\varnothing)$]{
        \centering
        \includegraphics[width=1.in, height=1.in]{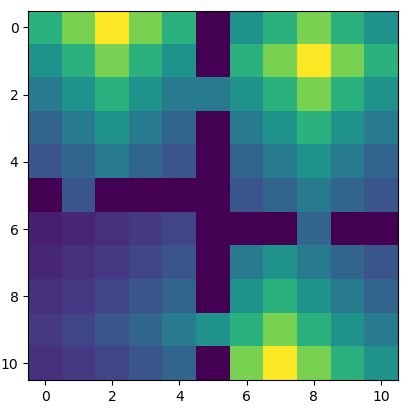}
        \label{fig:vis_room_0}
    }
    \hspace{5pt}
    \subfigure[$Q^{\theta}_C(\cdot,*;B)$]{
        \centering
        \includegraphics[width=1.in, height=1.in]{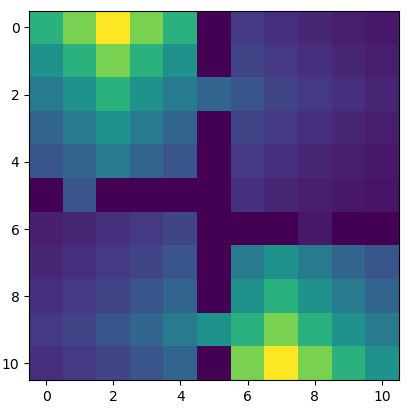}
        \label{fig:vis_room_1}
    }
    \hspace{5pt}
    \subfigure[$Q^{\theta}_C(\cdot,*;B,A)$]{
        \centering
        \includegraphics[width=1.in, height=1.in]{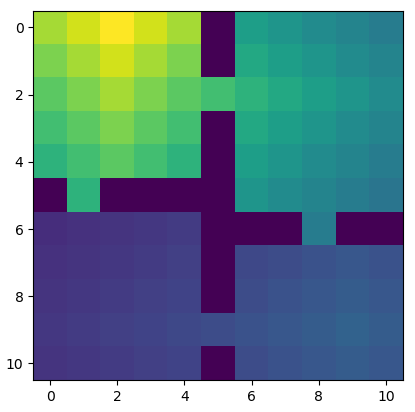}
        \label{fig:vis_room_2}
    }
    \caption{Visualization of trained action-value function of options. The first row is for the option of reaching $a$ in letter domain, and the second row is for the option of reaching $C$ in room domain. The color in every grid (state $s$) corresponds to the Q value of the optimal action, i.e., $\forall s, Q^{\theta}_p(s,*;\xi)=max_a Q^{\theta}_p(s,a;\xi)$. 
    }
    \label{fig:vis}
\end{figure*}

Finally, in order to show the effects of the dependence of options on future subgoals, we visualize the Q functions of the same option dependent on different future subgoals. The color of very grid represents the discounted return to the target subgoal, where the brighter the color is, the higher the return will be. 

In Figure \ref{fig:vis}, the first row shows the Q functions of reaching subgoal $a$ in letter domain dependent on nothing, $b$ and $b\to c$. Every grid represents the environment state where the agent is in that grid. On the map shown in Figure \ref{fig:vis_letter_env}, there are three different $a$, and according to Figure \ref{fig:vis_letter_0}, the agent should go to the closest $a$. The Figures \ref{fig:vis_letter_1} and \ref{fig:vis_letter_2} tell us that when dependent on $b$ ($b\to c$), the option of reaching $a$ regards the $a$ in first row (7-th row) as the target.

In the second row of Figure \ref{fig:vis}, we can see that in room domain, the option of reaching $C$ has different targets when the future subgoal sequences $\xi$ are different. Specifically, In Figure \ref{fig:vis_room_2}, the grid containing the yellow key has the highest value in the bottom rooms and the grid having $C$ in the upper left room has the highest value across the whole map. This indicates that in environment states where the agent is in the bottom rooms, the agent should first go to pick up the yellow key as an intermediate target and then go to $C$ in the upper left corner. It shows that the agent successfully learns the skill of opening a lock by the right key, without having any key proposition or prior knowledge.

\subsection{Neural Network Architecture}
\label{sec:arch}

\begin{wrapfigure}{l}{0.35\textwidth}
    \centering
    \includegraphics[width=2in]{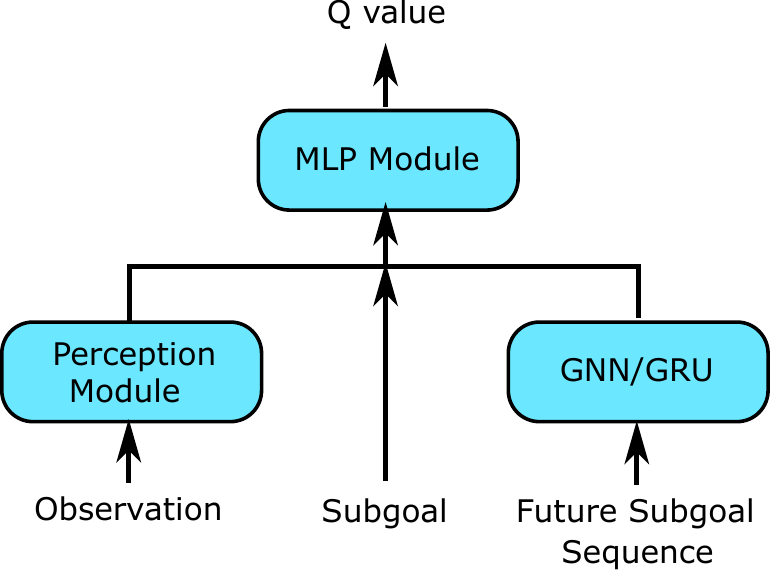}
    \caption{Neural Architecture of $Q^{\theta}(\cdot,\cdot;\xi)$}
    \label{fig:q_function}
\end{wrapfigure}

The agent's architecture of critic (Q function) $Q^{\theta}$ is shown in Figure \ref{fig:q_function}. The input consists of observation, subgoal embedding and subgoal sequence. The observation is processed by the perception module. The subgoal embedding is the one-hot encoding of the subgoal proposition. And the future subgoal sequence is processed by GNN or GRU. After inputs are processed, the embeddings of observation, subgoal and future subgoal sequence are concatenated and fed into an MLP to predict the return. The value function $V^{\phi}$ has the same architecture as Q function, except that its inputs only have observation and future subgoal sequence. 

The perception module is determined by the observation space of the environment: in letter/room domain with map size of $n\times n$, we used a 3-layer convolutional neural network (CNN) which have 16, 32 and 64 channels, respectively, where the kernel size is $l\in\{2, 3, 4\}$ and stride is 1; in navigation domain, we used a 2-layer fully-connected network with [256, 256] units and ReLU activations.

The sequence of future subgoal is processed by GNN or GRU here. The GNN used here is a graph convolutional network (GCN) \cite{kipf2016semi,schlichtkrull2018modeling} with 8 message passing steps and 32-dimensional node embeddings. The GRU used here is a 2-layer bidirectional GRU \cite{cho2014properties} with a 32-dimensional hidden layer.

For the MLP part of Q function in Figure \ref{fig:q_function}, we use 3 fully-connected layers with [64, 64, $d_a$] units and ReLU activations for all three domains. For discrete action space environments, $a_d$ is the number of possible actions, and the output of Q function was passed through a logit layer before softmax. For the continuous case, $a_d$ is the action dimension and we also need to train an actor network sharing same architecture as Q network except the Tanh activation. Then we assume a Gaussian action distribution and parameterized its mean and standard deviation by sending the actor’s output to two separate linear layers. 

In three baselines, the Q/value networks and actor network of the agent have the same architectures introduced here, keeping the same model complexity as the proposed method. In baseline-1, since the option does not consider future subgoals, the Q network does not have module to process subgoal sequence. In baseline-2 and 3, since they do not use options, the Q network and actor network do not have subgoal as input, where LTL formula is first transformed into a syntax tree and processed by a GCN (in baseline-2) or GRU (in baseline-3 without progression). The GCN in baseline-2 has the same architecture as that introduced above. The GRU in baseline-3 is also a 2-layer bidirectional GRU with 32-dimensional hidden layers.

\subsection{Algorithm Hyper-parameters}
\label{sec:hyper}
All experiments were conducted on a compute cluster using 1 GPU (RTX 2080 Ti). The hyper-parameters used for deep Q learning in letter and room domain are introduced in Table \ref{tab:hyp_letter} and Table \ref{tab:hyp_room}. The hyper-parameters for SAC in navigation domain are presented in Table \ref{tab:hyp_navi}. 

The agents in three baselines are trained by the same RL algorithms in the proposed method, using the same algorithm hyperparameters of the proposed method. In baseline-1, we do not consider future subgoal sequence. In baseline-2 and 3, we cannot use LTL decomposition or HER since the LTL formula is transformed into a syntax tree from its original form. 

\begin{table}[ht]
\centering
\caption{Hyperparameters of Deep Q Learning in Letter Domain}
\label{tab:hyp_letter}
\begin{tabular}{cc}
\hline
\hline
{\bf Hyperparameter}     & {\bf Value} \\\hline
 Batch size & 256 \\
 Discount & 0.99 \\
 Exploration $\epsilon$ init value & 0.75 \\
 Exploration $\epsilon$ final value & 0.05 \\
 Exploration $\epsilon$ factor & 0.5 \\
 Curriculum level $K$ & 5 \\
 Total number of steps & 10e6 \\
 Satisfaction Reward $R_F$ & 10 \\
 Step penalty $R_e$ & -0.01 \\
 $Q$ update interval & 10000 \\
 $Q$ target update interval & 2000 \\
 $V$ update interval & 10 \\
 $V$ target update interval & 2000 \\
 HER trajectory modification ratio & 0.5 \\
 Evaluation interval & 10 \\
 Evaluation episodes & 10 \\
 Optimizer & Adam \\
 Adam $\epsilon$ & $2\times 10^{-5}$ \\
 $\beta_1, \beta_2$ & $0.9, 0.999$ \\
 Learning rate & $3\times 10^{-4}$ \\
 Replay buffer size $|\mathcal{B}|$ & 2e6 
\end{tabular}
\end{table}

\begin{table}[ht]
\centering
\caption{Hyperparameters of Deep Q Learning in Room Domain}
\label{tab:hyp_room}
\begin{tabular}{cc}
\hline
\hline
{\bf Hyperparameter}     & {\bf Value} \\\hline
 Batch size & 256 \\
 Discount & 0.99 \\
 Exploration $\epsilon$ init value & 0.75 \\
 Exploration $\epsilon$ final value & 0.05 \\
 Exploration $\epsilon$ factor & 0.5 \\
 Curriculum level $K$ & 5 \\
 Total number of steps & 10e6 \\
 Step penalty $R_e$ & -0.01 \\
 Satisfaction Reward $R_F$ & 10 \\
 $Q$ update interval & 5 \\
 $Q$ target update interval & 1500 \\
 $V$ update interval & 5 \\
 $V$ target update interval & 1500 \\
 HER trajectory modification ratio & 1.0 \\
 Evaluation interval & 10000 \\
 Evaluation episodes & 10 \\
 Optimizer & Adam \\
 Adam $\epsilon$ & $2\times 10^{-5}$ \\
 $\beta_1, \beta_2$ & $0.9, 0.999$ \\
 Learning rate & $3\times 10^{-4}$ \\
 Replay buffer size $|\mathcal{B}|$ & 2e6 
\end{tabular}
\end{table}

\begin{table}[ht]
\centering
\caption{Hyperparameters of SAC in Navigation Domain}
\label{tab:hyp_navi}
\begin{tabular}{cc}
\hline
\hline
{\bf Hyperparameter}     & {\bf Value} \\\hline
 Batch size & 256 \\
 Discount & 0.995 \\
 Time limit in an episode & 1000 \\
 $\alpha$ schedule & automatic \\
 Curriculum level $K$ & 5 \\
 Total number of steps & 10e6 \\
 Satisfaction Reward $R_F$ & 10 \\
 $Q$ update interval & 5 \\
 $Q$ target update interval & 10000 \\
 $V$ update interval & 5 \\
 $V$ target update interval & 10000 \\
 HER trajectory modification ratio & 1.0 \\
 Evaluation interval (episodes) & 10 \\
 Evaluation episodes & 10 \\
 Optimizer & Adam \\
 Adam $\epsilon$ & $2\times 10^{-5}$ \\
 $\beta_1, \beta_2$ & $0.9, 0.999$ \\
 Learning rate & $2\times 10^{-4}$ \\
 Replay buffer size $|\mathcal{B}|$ & 1e6 
\end{tabular}
\end{table}

\end{document}